\documentclass[runningheads]{llncs}
\usepackage{graphicx}

\usepackage{tikz}
\usepackage{comment}
\usepackage{amsmath,amssymb} 
\usepackage{color}
\usepackage[colorlinks,
            linkcolor=blue,
            anchorcolor=blue,
            citecolor=blue]{hyperref}
            
\usepackage{booktabs}
\usepackage{array}
\usepackage{multirow}
\usepackage{listings,xcolor}
\usepackage{subcaption}
\usepackage{appendix}
\usepackage{makecell}
\newcommand{\mypm}[1]{\color{black}{\tiny{#1}}}
\newcommand{\mypmRED}[1]{\color{red}{{#1}}}
\newcommand{\mypmBLUE}[1]{\color{blue}{{#1}}}

\usepackage[accsupp]{axessibility}  

\begin{document}
\pagestyle{headings}
\mainmatter
\def\ECCVSubNumber{103}  

\title{PixelFolder: An Efficient Progressive Pixel Synthesis Network for Image Generation} 

\titlerunning{PixelFolder: An Efficient Progressive Pixel Synthesis Network}
%
\author{
Jing He$^{1}$ \and
Yiyi Zhou$^{1}$\thanks{~Corresponding Author. } \and
Qi Zhang$^{2}$ \and
Jun Peng$^{1}$ \and
Yunhang Shen$^{2}$\and
Xiaoshuai Sun$^{1,3}$ \and
Chao Chen$^{2}$ \and
Rongrong Ji$^{1,3}$
}
\authorrunning{J. He et al.}
%
\institute{
$^{1}$Media Analytics and Computing Lab, Department of Artificial Intelligence, School of Informatics, Xiamen University. $^{2}$Youtu Lab, Tencent.
~$^{3}$Institute of Artificial Intelligence, Xiamen University. \\
\email{\{blinghe, pengjun\}@stu.xmu.edu.cn},
\email{\{zhouyiyi, xssun, rrji\}@xmu.edu.cn},
\email{\{merazhang, aaronccchen\}@tencent.com},
\email{shenyunhang01@gmail.com}
}

\maketitle
\setcounter{footnote}{0}
\begin{abstract}

Pixel synthesis is a promising research paradigm for image generation, which can well exploit pixel-wise prior knowledge for generation. 
However, existing methods still suffer from excessive memory footprint and computation overhead. 
In this paper, we propose a progressive pixel synthesis network towards efficient image generation, coined as \emph{PixelFolder}. 
Specifically, PixelFolder formulates image generation as a progressive pixel regression problem and synthesizes images by a multi-stage paradigm, which can greatly reduce the overhead caused by large tensor transformations. 
In addition, we introduce novel \emph{pixel folding} operations to further improve model efficiency while maintaining pixel-wise prior knowledge for end-to-end regression. 
With these innovative designs, we greatly reduce the expenditure of pixel synthesis, \emph{e.g.}, reducing $89\%$ computation and $53\%$ parameters compared to the latest pixel synthesis method called \emph{CIPS}. 
To validate our approach, we conduct extensive experiments on two benchmark datasets, namely FFHQ and LSUN Church. The experimental results show that with much less expenditure, 
PixelFolder obtains new state-of-the-art (SOTA) performance on two benchmark datasets, \emph{i.e.}, $3.77$ \emph{FID} and $2.45$ \emph{FID} on FFHQ and LSUN Church, respectively.
Meanwhile, PixelFolder is also more efficient than the SOTA methods like \emph{StyleGAN2}, reducing about $72\%$ computation and $31\%$ parameters, respectively. 
These results greatly validate the effectiveness of the proposed PixelFolder. Our source code is available at \url{https://github.com/BlingHe/PixelFolder}. 

\end{abstract}

\begin{figure}
    \centering
    \subcaptionbox{
    straight hair \label{subfig:sh}}{
    \includegraphics[width =.475\linewidth]{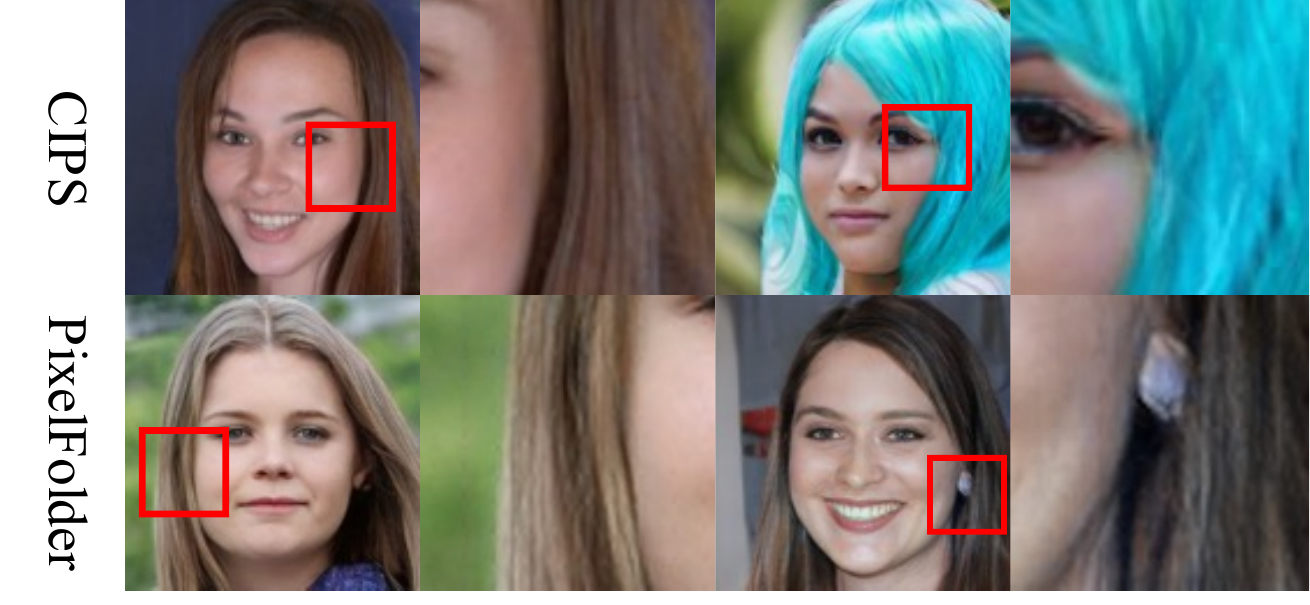}
    }
    \hfill
    \subcaptionbox{wavy hairs \label{subfig:wh}}{
    \includegraphics[width = .475\linewidth]{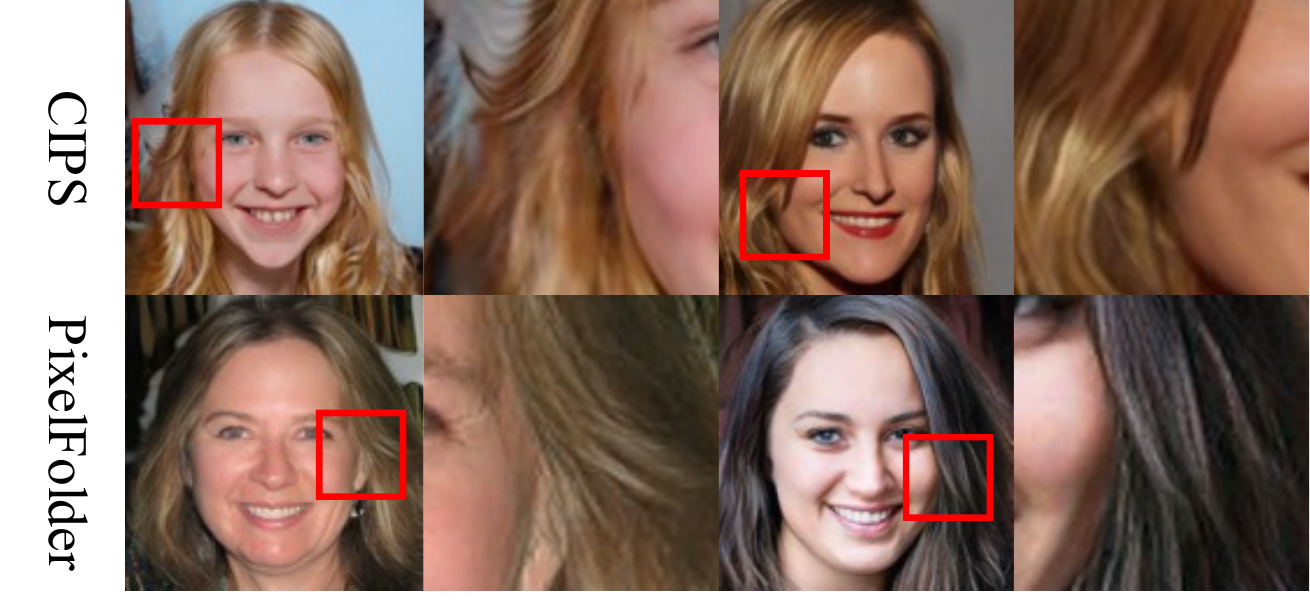}
    }
    \hfill
    \subcaptionbox{bangs \label{subfig:b}}{
    \includegraphics[width = .475\linewidth]{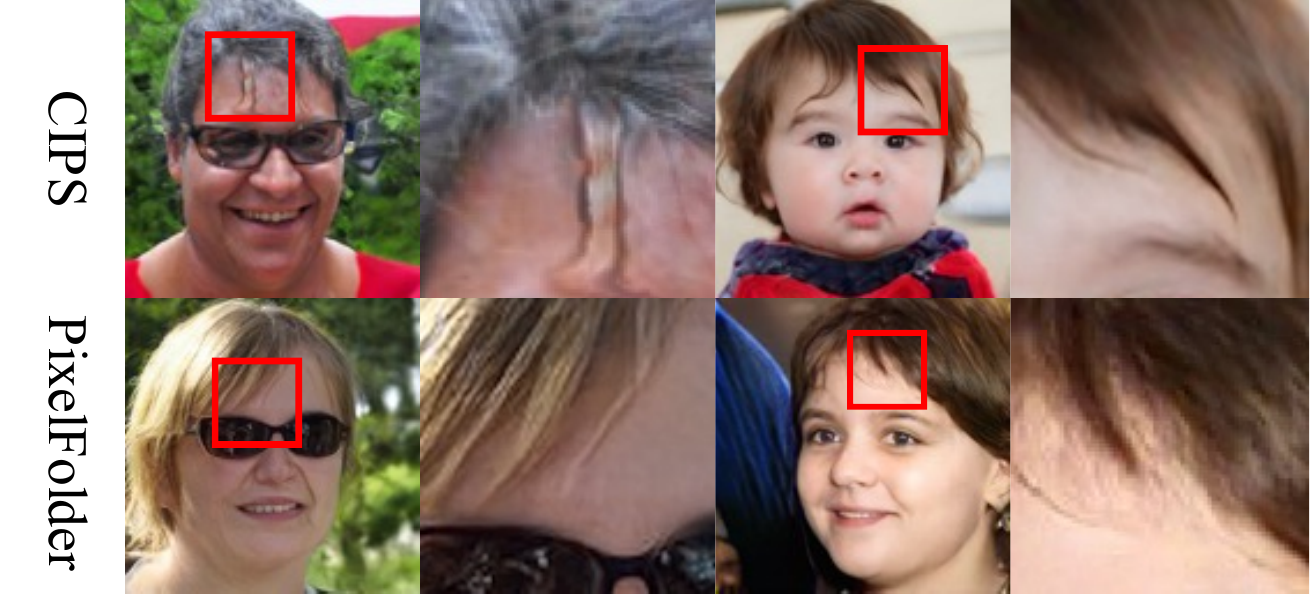}
    }\hfill
    \subcaptionbox{short hairs \label{subfig:shth}}{
    \includegraphics[width = .475\linewidth]{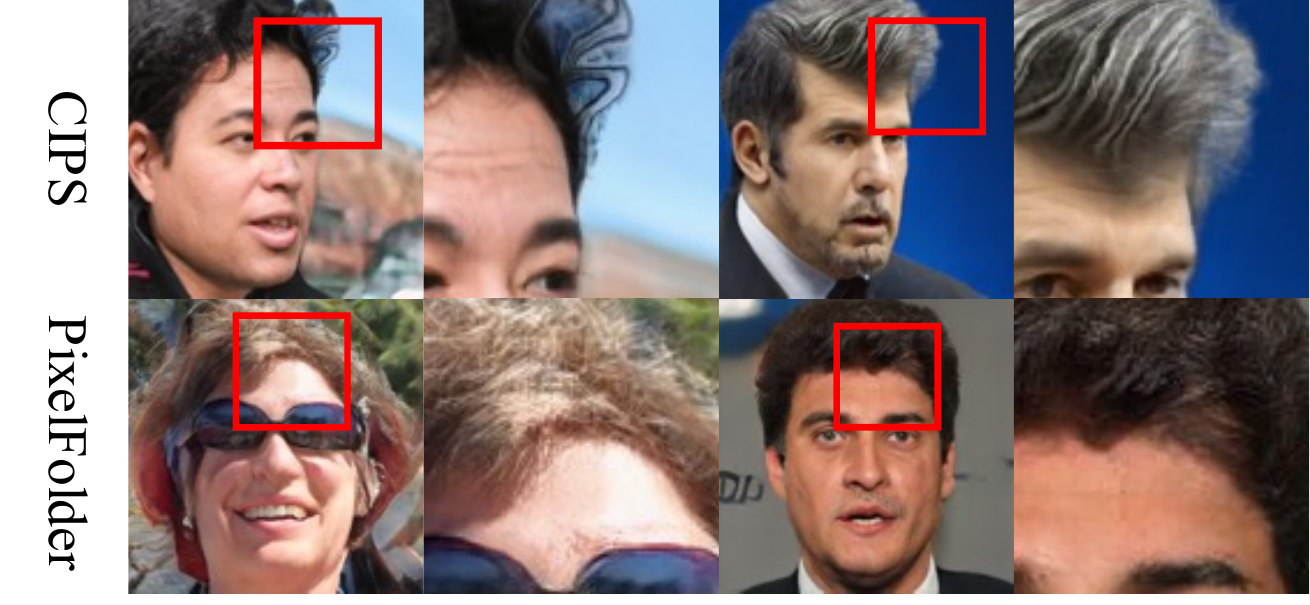}
    }

\caption{
Comparison of the generated faces by CIPS~\cite{anokhin2021image} and PixelFolder on FFHQ. Compared with CIPS, PixelFolder synthesizes more vivid faces and can also alleviate local incongruities via its novel network structure. 
}

\label{fig:local_incongruities}
\end{figure}

\section{Introduction}
As an important task of computer vision, image generation has made remarkable progress in recent years, which is supported by a flurry of generative adversarial networks~\cite{brock2018large,chen2016infogan,denton2015deep,goodfellow2014generative,hudson2021generative,karras2018progressive,karras2019style,karras2020analyzing,lee2021vitgan,radford2015unsupervised}.
One of the milestone works is the StyleGAN series~\cite{karras2019style,karras2020analyzing}, which borrows the principle of style transfer~\cite{huang2017arbitrary} to build an effective generator architecture. Due to the superior performance in image quality, this style-driven modeling has become the mainstream paradigm of image generation~\cite{karras2019style,karras2020analyzing}, which also greatly influences and promotes the development of other generative tasks, such as image manipulation~\cite{dolhansky2018eye,kim2021exploiting,wang2021hififace,yu2018generative,zhu2020sean}, image-to-image translation~\cite{choi2018stargan,isola2017image,ji2022knowing,li2021image,ma2022knowing,zhu2017unpaired} and text-to-image generation~\cite{li2019controllable,park2020swapping,peng2021knowledge,xu2018attngan}.

In addition to the StyleGAN series, pixel synthesis~\cite{anokhin2021image,skorokhodov2021adversarial} is another paradigm of great potential for image generation. Recently, Anokin \emph{et al.}~\cite{anokhin2021image} propose a novel Conditionally-Independent Pixel Synthesis (CIPS) network for adversarial image generation, which directly computes each pixel value based on the random latent vector and positional embeddings. This end-to-end pixel regression strategy can well exploit pixel-wise prior knowledge to facilitate the generation of high-quality images. Meanwhile, it also simplifies the design of generator architecture, \emph{e.g.}, only using $1\times 1$ convolutions, and has a higher 
generation ability with non-trivial topologies~\cite{anokhin2021image}. On multiple benchmarks \cite{karras2019style,radford2015unsupervised}, this method exhibits comparable performance against the StyleGAN series, showing a great potential in image generation. In this paper, we also follow the principle of pixel synthesis to build an effective image generation network.

\begin{figure}[!h]
\centering
\subcaptionbox{CIPS \label{subfig:strc_cips}}{
\includegraphics[width =.17\linewidth]{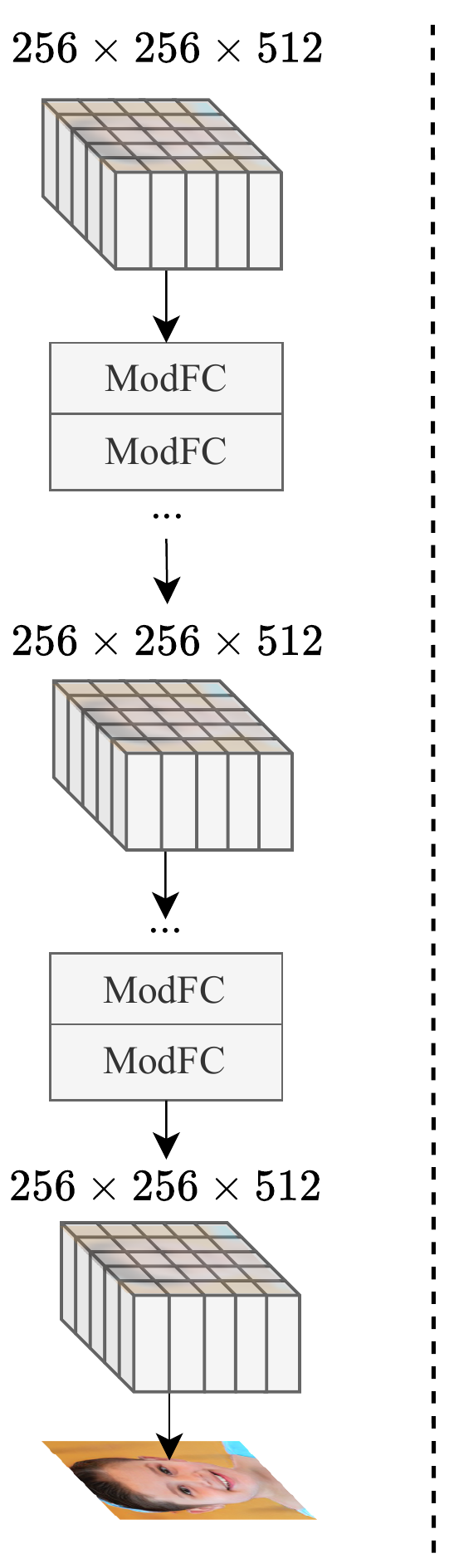}
}
\hfill
\subcaptionbox{PixelFolder (Ours) \label{subfig:pxf}}{
\includegraphics[width = .77\linewidth]{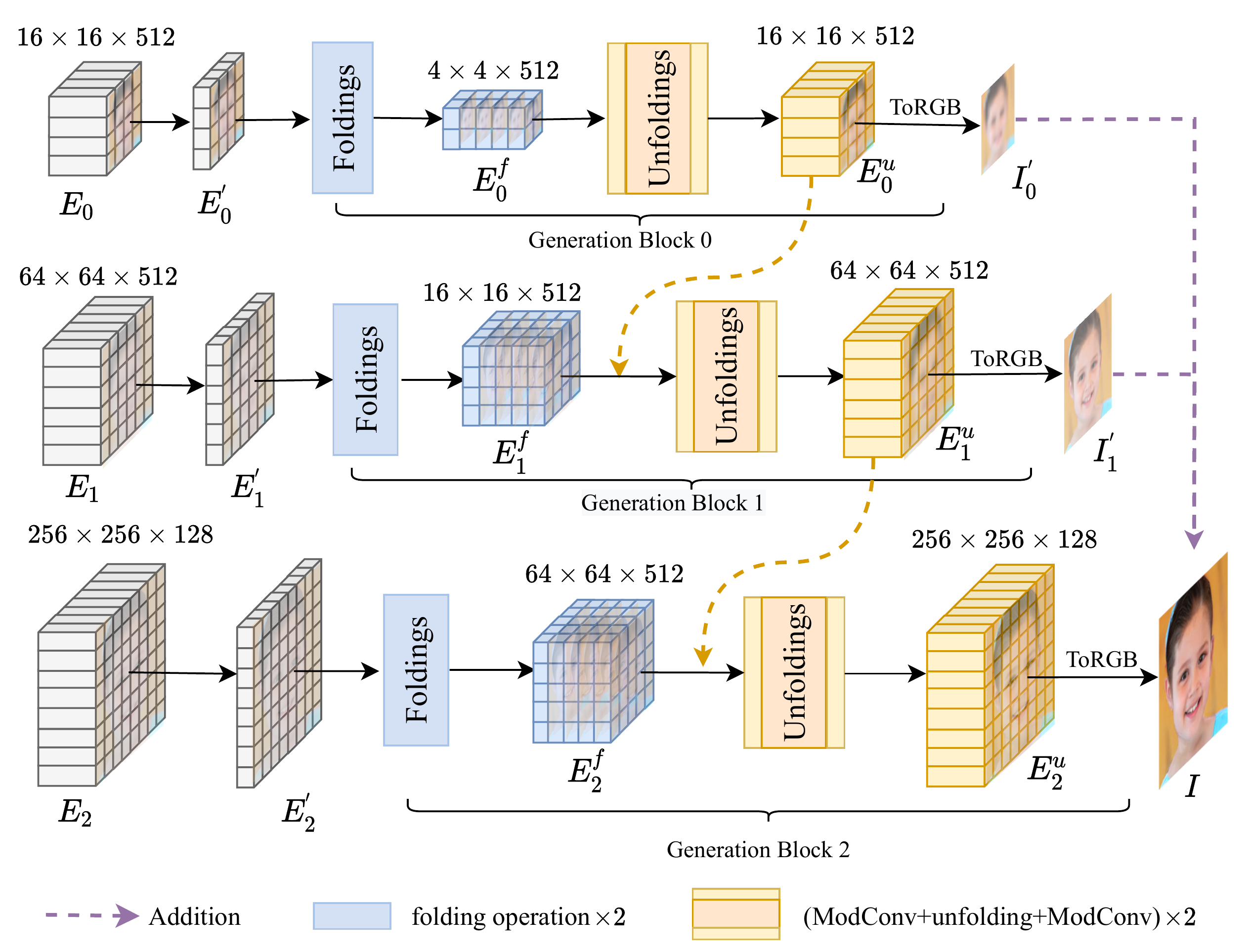}}
\caption{
A comparison of the architectures of CIPS~\cite{anokhin2021image} (left) and the proposed PixelFolder (right). 
PixelFolder follows the pixel synthesis principle of CIPS, but regards image generation as a multi-stage regression problem, thereby reducing the cost of large tensor transformations. Meanwhile, novel \emph{pixel folding} operations are also applied in PixelFodler to further improve model efficiency. 
}
\label{fig:structure}
\end{figure}

Despite the aforementioned merits, CIPS still has obvious shortcomings in model efficiency. Firstly, although CIPS is built with a simple network structure, it still requires excessive memory footprint and computation during inference. Specifically, this is mainly attributed to its high-resolution pixel tensors for end-to-end pixel regression, \emph{e.g.}, $256\times 256\times 512 $, which results in a large computational overhead and memory footprint, as shown in Fig.~\ref{subfig:strc_cips}. 
Meanwhile, the learnable coordinate embeddings also constitute a large number of parameters, making CIPS taking about $30\%$ more parameters than StyleGAN2~\cite{karras2020analyzing}. 
These issues greatly limit the applications of CIPS in high-resolution image synthesis. 

To address these issues, we propose a novel progressive pixel synthesis network towards efficient image generation, termed \emph{PixelFolder}, of which structure is illustrated in Fig.~\ref{subfig:pxf}. 
Firstly, we transform the pixel synthesis problem to a progressive one and then compute pixel values via a multi-stage structure. 
In this way, the generator can process the pixel tensors of varying scales instead of the fixed high-resolution ones, thereby reducing memory footprint and computation greatly. 
Secondly, we introduce novel \emph{pixel folding} operations to further improve model efficiency.
In PixelFolder, the large pixel tensors of different stages are folded into the smaller ones, and then gradually unfolded (expanded) during feature transformations. These pixel folding (and unfolding) operations can well preserve the independence of each pixel, while saving model expenditure. These innovative designs help PixelFolder achieves high-quality image generations with superior model efficiency, which are also shown to be effective for \emph{local imaging incongruity} found in CIPS~\cite{anokhin2021image}, as shown in Fig.~\ref{fig:local_incongruities}.

To validate the proposed PixelFolder, we conduct extensive experiments on two benchmark datasets of image generation, \emph{i.e.}, FFHQ~\cite{karras2019style} and LSUN Church~\cite{radford2015unsupervised}.
The experimental results show that PixelFolder not only outperforms CIPS in terms of image quality on both benchmarks, but also reduces parameters and computation by $53\%$ and $89\%$, respectively.
Compared to the state-of-the-art model, \emph{i.e.}, StyleGAN2~\cite{karras2020analyzing}, PixelFolder is also very competitive and obtains new SOTA performance on FFHQ and LSUN Church, \emph{i.e.}, $3.77$ FID and $2.45$ FID, respectively.
Meanwhile, the efficiency of PixelFolder is still superior, with $31\%$ less parameters and $72\%$ less computation than StyleGAN2. 

To sum up, our contribution is two-fold:
\begin{enumerate}
    \item We propose a progressive pixel synthesis network for efficient image generation, termed \emph{PixelFolder}. With the multi-stage structure and innovative pixel folding operations, PixelFolder greatly reduces the computational and memory overhead while keeping the property of end-to-end pixel synthesis. 
    \item Retaining much higher efficiency, the proposed PixelFolder not only has better performance than the latest pixel synthesis method CIPS, but also achieves new SOTA performance on FFHQ and LSUN Church. 
\end{enumerate}

\section{Related Work}
Recent years have witnessed the rapid development of image generation supported by a bunch of generative adversarial network (GAN) \cite{goodfellow2014generative} based methods~\cite{afifi2021histogan,liang2021high,lin2021anycost,liu2021divco,he2021eigengan,park2020contrastive,patashnik2021styleclip,tang2020xinggan,wang2021image}.
Compared with previous approaches~\cite{kingma2013auto,van2017neural}, GAN-based methods model the domain-specific data distributions better through the specific adversarial training paradigm, \emph{i.e.}, a discriminator is trained to distinguish whether the images are true or false for the optimization of the generator. To further improve the quality of generations, a flurry of methods~\cite{denton2015deep,radford2015unsupervised,chen2016infogan,arjovsky2017wasserstein,gulrajani2017improved} have made great improvements in both GAN structures and objective functions. 
Recent advances also resort to a progressive structure for high-resolution image generation. 
PGGAN~\cite{karras2018progressive} proposes a progressive network to generate high-resolution images, where both generator and discriminator start their training with low-resolution images and gradually increase the model depth by adding-up the new layers during training.
StyleGAN series~\cite{karras2019style,karras2020analyzing} further borrow the concept of ``\emph{style}'' into the image generation and achieve remarkable progress. The common characteristic of these progressive methods is to increase the resolution of hidden features by up-sampling or deconvolution operations. Differing from these methods, our progressive modeling is based on the principle of pixel synthesis with pixel-wise independence for end-to-end regression.

In addition to being controlled by noise alone, some methods exploit coordinate information for image generation. 
CoordConv-GAN~\cite{liu2018intriguing} introduces pixel coordinates in every convolution based on DCGAN~\cite{radford2015unsupervised}, which proves that pixel coordinates can better establish geometric correlations between the generated pixels. COCO-GAN~\cite{lin2019coco} divides the image into multiple patches with different coordinates, which are further synthesized independently. CIPS~\cite{anokhin2021image} builds a new paradigm of using coordinates for image generation, \emph{i.e.}, pixel regression, which initializes the prior matrix based on pixel coordinates and deploys multiple $1 \times 1$ convolutions for pixel transformation. This approach not only greatly simplifies the structure of generator, but also achieves competitive performance against existing methods. In this paper, we also follow the principle of pixel regression to build the proposed PixelFolder. 

Our work is also similar to a recently proposed method called INR-GAN~\cite{skorokhodov2021adversarial}, which also adopts a multi-stage structure. In addition to the obvious differences in network designs and settings, PixelFolder is also different from INR-GAN in the process of pixel synthesis. In INR-GAN, the embeddings of pixels are gradually up-sampled via \emph{nearest neighbor interpolation}, which is more in line with the progressive models like StyleGAN2~\cite{karras2020analyzing} or PGGAN~\cite{karras2018progressive}. In contrast, PixelFolder can well maintain the independence of each pixel during multi-stage generation, and preserve the  property of end-to-end pixel regression via pixel folding operations.

\section{Preliminary}
\label{sec:pre}
Conditionally-Independent Pixel Synthesis (CIPS) is a novel generative adversarial network proposed by Anokhin \emph{et al.}~\cite{anokhin2021image}. Its main principle is to synthesis each pixel conditioned on a random vector $z\in Z$ and the pixel coordinates $(x,y)$, which can be defined by
\begin{equation}
\label{eq:cips}
    I = \left \{G(x,y;\mathbf{z})|(x,y) \in mgrid(H,W) \right \},
\end{equation}
where $mgrid(H,W)=\left \{ (x,y)|0\leq x \leq W, 0 \leq y \leq H \right \}$ is the set of integer pixel coordinates, and $G(\cdot)$ is the generator. 
Similar to StyleGAN2 \cite{karras2020analyzing}, $z$ is turned into a style vector $w$ via a mapping network and then shared by all pixels. Afterwards, $w$ is injected into the generation process via ModFC layers \cite{anokhin2021image}. 

An important design in CIPS is the positional embeddings of synthesized pixels, which are consisted of Fourier features and coordinate embeddings. The Fourier feature of each pixel $e_{fo}(x,y)\in\mathbb{R}^d$ is computed based on the coordinate $(x,y)$ and transformed by a learnable weight matrix $B_{fo}\in \mathbb{R}^{2\times d}$ and $sin$ activation. 
To improve  model capacity, Anokhin \emph{et al.} also adopt the coordinate embedding $e_{co}(x,y)\in\mathbb{R}^d$ , which has $H\times W$ learnable vectors in total. 
Afterwards, the final pixel vector $e(x,y) \in\mathbb{R}^{d}$ is initialized by concatenating these two types of embeddings and then fed to the generator. 

Although CIPS has a simple structure and can be processed in parallel \cite{anokhin2021image}, its computational cost and memory footprint are still expensive, mainly due to the high-resolution pixel tensor for end-to-end generation. In this paper, we follow the principle of CIPS defined in Eq.~\ref{eq:cips} to build our model and address the issue of model efficiency via a progressive regression paradigm.

\section{PixelFolder}
\label{sec:pxf}
\subsection{Overview}
\label{subsec:overview}

The structure of the proposed PixelFodler is illustrated in Fig.\ref{fig:structure}. To reduce the high expenditure caused by end-to-end regression for large pixel tensors, we first transform pixel synthesis to a multi-stage generation problem, which can be formulated as
\begin{equation}
    I=\sum_{i=0}^{K-1} \left \{ G_i(x_i,y_i;\mathbf{z})|(x_i,y_i)\in mgrid(H_i,W_i) \right \},
\end{equation}
where $i$ denotes the index of generation stages.
At each stage, we initialize a pixel tensor $\mathbf{E}_i\in \mathbb{R}^{H_i\times W_i \times d}$ for generation. The \emph{RGB} tensors  ${I'}_i\in\mathbb{R}^{H_i\times W_i\times 3}$ predicted by different stages are then aggregated for the final pixel regression. This progressive paradigm can avoid the constant use of large pixel tensors to reduce excessive memory footprint.  In literature \cite{karras2018progressive,skorokhodov2021adversarial,zhang2018stackgan++,zhang2018photographic}, it is also shown effective to reduce the difficulty of image generation.

To further reduce the expenditure of each generation stage, we introduce novel \emph{pixel folding} operations to PixelFolder. As shown in Fig.\ref{fig:structure}, the large pixel tensor is first projected onto a lower-dimension space, and their local pixels, \emph{e.g.}, in $2\times 2$ patch, are then concatenated to form a new tensor with a smaller resolution, denoted as $\mathbf{E}_i^{f}\in\mathbb{R}^{\frac{H_i}{k}\times \frac{W_i}{k}\times d}$, where $k$ is the scale of folding. After passing through the convolution layers, the pixel tensor is decomposed again (truncated from the feature dimension), and combined back to the original resolution. We term these parameter-free operations as \emph{pixel folding} (and unfolding). Folding features is not uncommon in computer vision, which is often used as an alternative to the operations like \emph{down-sampling} or \emph{pooling}~\cite{liu2018deep,luo2021towards,luo2022towards,shi2016real}. But in PixelFolder, it not only acts to reduce the tensor resolution, but also serves to maintain the independence of folded pixels. 

To maximize the use of pixel-wise prior knowledge at different scales, we further combine the folded tensor $E_i^{f}$ with the unfolded pixel tensor $E_{i-1}^{u}$ of the previous stage, as shown in Fig.~\ref{subfig:pxf}.
With the aforementioned designs, PixelFolder can significantly reduce memory footprint and computation, while maintaining the property of pixel synthesis.

\begin{figure}[!t]
\centering
\subcaptionbox{pixel foldings \label{subfig:foldings}}{
\centering
\includegraphics[width =.36\linewidth]{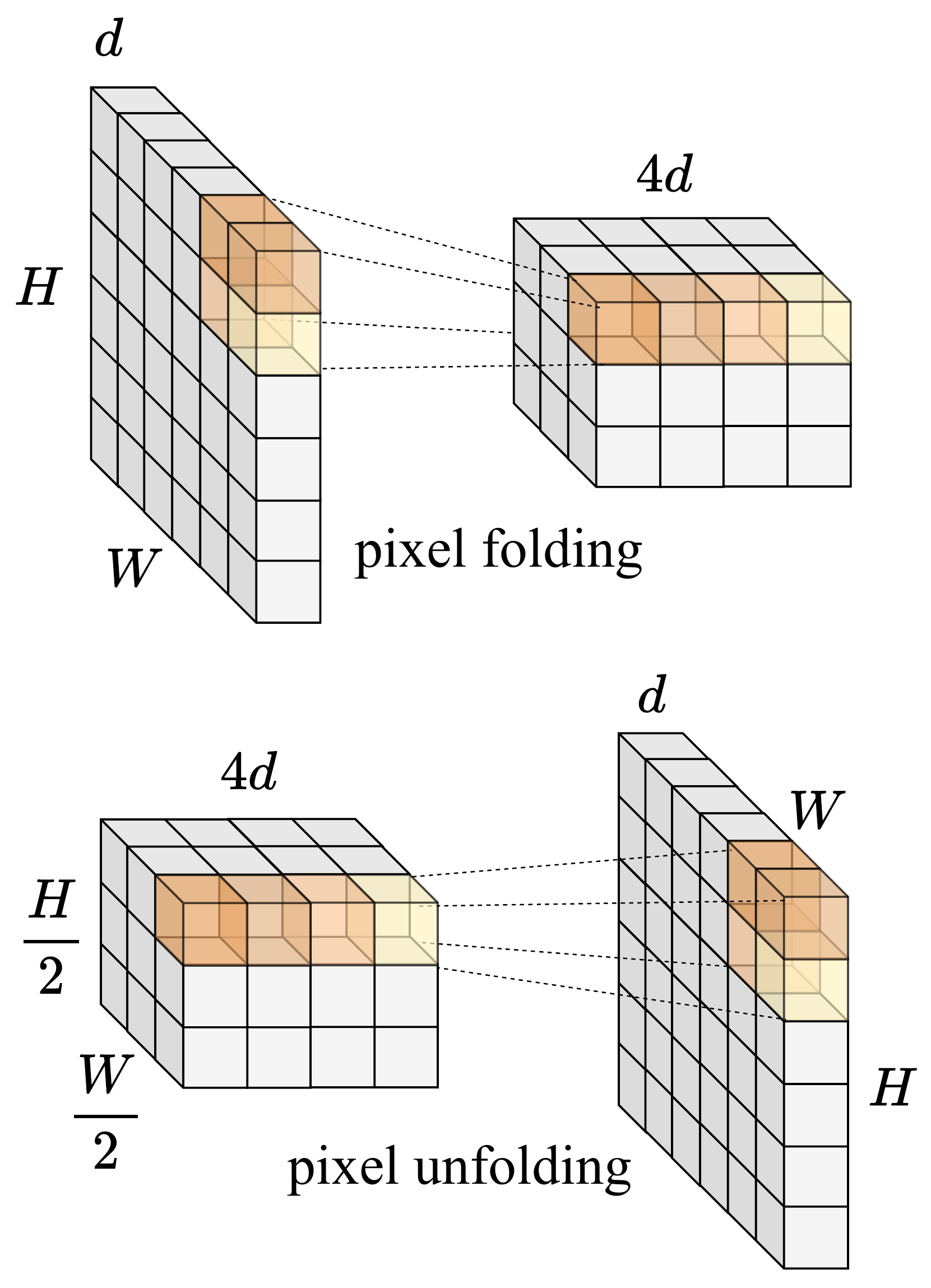}
}
\hfill
\subcaptionbox{generation block \label{subfig:generation_block}}{
\includegraphics[width = .58\linewidth]{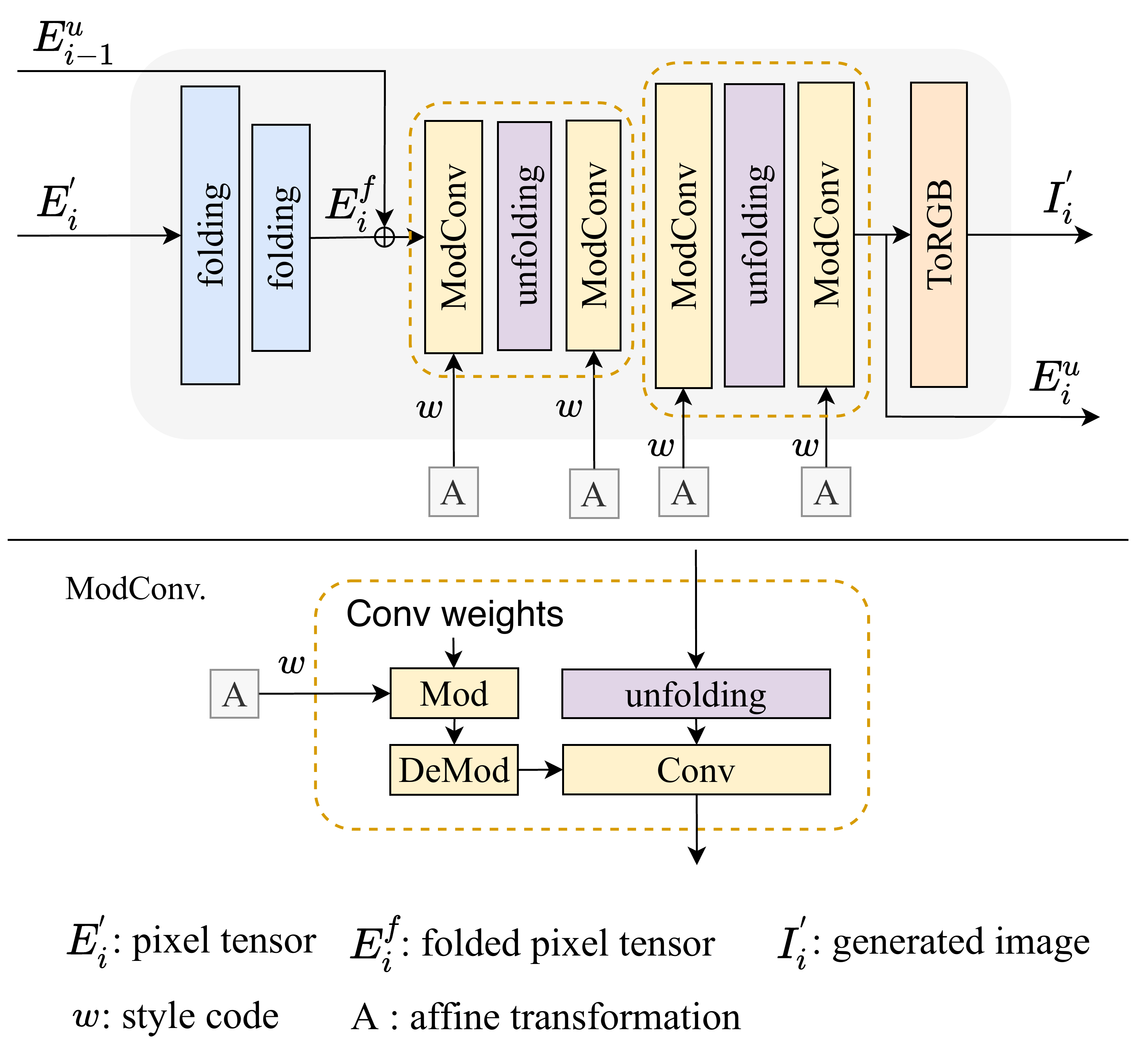}
}

\caption{
(a) The illustrations of pixel folding and unfolding operations. These parameter-free operations can maintain the pixel-wise independence when changing the tensor resolution. (b) The detailed structure of the generation block in PixelFolder. The number of parameterized layers in PixelFolder is much smaller than those of CIPS and StyleGAN2. 
}
\label{fig:folding}
\end{figure}

\subsection{Pixel folding}

The illustration of \emph{pixel folding} is depicted in Fig.~\ref{subfig:foldings}, which consists of two operations, namely \emph{folding} and \emph{unfolding}. 
The folding operation spatially decomposes the pixel tensor into multiple local patches, 
and straighten each of the patches to form a smaller but deeper tensor. 
On the contrary, the unfolding operation will truncate the folded pixel vectors from the feature dimension to recover the tensor resolution.

 Particularly, pixel folding can effectively keep the independence and spatial information of each pixel regardless of the varying resolutions of the hidden tensors. This also enables the pixel-wise prior knowledge to be fully exploited for image generation. In addition, when the pixels are folded, they can receive more interactions via convolutions, which is found to be effective for the issue of \emph{local imagery incongruity} caused by insufficient local modeling~\cite{anokhin2021image}.

\subsection{Pixel tensor initialization }
\label{subsec:init_tensor}
Similar to CIPS \cite{anokhin2021image}, we also apply Fourier features and coordinate embeddings to initialize the pixel tensors. Specifically, given the coordinate of a pixel $(x,y)$, Fourier feature $e_{fo}(x,y)$ is obtained by
\begin{equation}
    e_{fo}(x,y) = \sin \left [ B_{fo}{(x',y')}^T \right ],
\end{equation}
where $x'=\frac{2x}{W_i-1}-1$ and $y'=\frac{2y}{H_i-1}-1$, and $B_{fo}\in \mathbb{R}^{2\times d}$ is the projection weight matrix. The coordinate embedding is a parameterized vector, denoted as $e_{co}(x,y)\in\mathbb{R}^{d}$. 
Afterwards, these two types of embeddings are concatenated and projected to obtain the new pixel tensor, denoted as  $\mathbf{E}_i\in\mathbb{R}^{H_i\times W_i\times d}$.

In principle, Fourier features serve to preserve the spatial information and capture the relationships between pixels~\cite{anokhin2021image,liu2018intriguing}. The learnable coordinate embeddings can increase model capacity to improve image quality, \emph{e.g.}, to avoid wave-like artifacts~\cite{anokhin2021image}. In  PixelFolder, we only apply coordinate embeddings to the first generation stage to keep model compactness, and we found this trade-off has little detriment to image quality during experiments.

\subsection{Generation blocks}
\label{subsec:genblk}
The detailed structure of generation blocks in PixelFolder is given in Fig.~\ref{subfig:generation_block}. After folding operations, a modulated convolution (\emph{ModConv}) layer~\cite{karras2020analyzing} is deployed for feature transformation. Then unfolding operations are used to recover the resolution, each followed by another ModConv layer. In practice, we use two folding and unfolding operations to gradually reduce and recover the tensor resolution, respectively, which is to avoid the drastic change of tensor resolution during feature transformation. The convolution filter is set to $3\times 3$, considering the issue of local imaging incongruity. 
Besides, we also carefully set the resolution and folded pixels of each generation stage to ensure that the output tensor of current stage can be integrated into the next stage. 
Similar to StyleGAN2~\cite{karras2020analyzing}, the style vector $w$ is injected into the ModConv layers via modulating their convolution filter, \emph{i.e.}, being mapped to scale vector $s$ with an affine network. Finally, the recovered pixel tensors are linearly projected onto RGB space as the output of each stage, which are then aggregated for the final regression. Due to our efficient modeling strategy, PixelFolder uses only 12 convolution layers in all generation stages, thus having much fewer parameters than the SOTA methods like StyleGAN2~\cite{karras2020analyzing} and CIPS~\cite{anokhin2021image}.

\section{Experiments}
To validate the proposed PixelFolder, we conduct extensive experiments on two benchmark datasets\footnote{More experiments on other datasets and high-resolution are available in the supplementary material.}, namely Flickr Faces-HQ  \cite{karras2019style} and LSUN Church \cite{radford2015unsupervised}, and compare it with a set of  state-of-the-art (SOTA) methods including CIPS \cite{anokhin2021image}, StyleGAN2~\cite{karras2020analyzing} and INR-GAN~\cite{skorokhodov2021adversarial}.  

\subsection{Datasets}

\textbf{Flickr Faces-HQ}~(FFHQ)~\cite{karras2019style} consistes of $70,000$ high-quality human face images, which all have a resolution of $1024 \times 1024$.
The images were crawled from Flickr and automatically aligned and cropped.

\noindent\textbf{LSUN Church} is the sub-dataset of Large-scale Scene UNderstanding(LSUN) benchmark~\cite{radford2015unsupervised}. 
It contains about $126,000$ images of churches in various architectural styles, which are collected from natural surroundings. 

\subsection{Metrics}
To validate the proposed PixelFolder, we conduct evaluations from the aspects of image quality and model efficiency, respectively.
The metrics used for image quality include \emph{Fr\'echet Inception Distance} (FID)~\cite{heusel2017gans} and \emph{Precision and Recall} (P\&R)~\cite{kynkaanniemi2019improved,sajjadi2018assessing}. FID measures the distance between the real images and the generated  ones from the perspective of mean and covariance matrix. 
P\&R evaluates the ability of fitting the true data distribution.
Specifically, for each method, we randomly generate $50,000$ images for evaluation. 
In terms of model efficiency, we adopt the number of parameters (\#Params), \emph{Giga Multiply Accumulate Operations} (GMACs)~\cite{howard2017mobilenets}, and generation speed (im/s) to measure model compactness, computation overhead and model inference, respectively.

\subsection{Implementation}
In terms of the generation network, we deploy three generation stages for PixelFolder, and their resolutions are set to 16, 64 and 256, respectively.
In these operations, 
the scale of folding and unfolding $k$ is set to $2$, \emph{i.e.}, the size of local patches is $2\times2$. 
The dimensions of initialized tensors are all 512, except for the last stage which is set to 128. Then these initialized tensors are all reduced to 32 via linear projections before pixel folding. The recovered pixel tensors after pixel unfolding are also projected to \emph{RGB} by linear projections. 
For the discriminator, we use a residual convolution network  following the settings in StyleGAN2 \cite{karras2020analyzing} and CIPS \cite{anokhin2021image}, which has \emph{FusedLeakyReLU}
activation functions and minibatch standard deviation layers~\cite{karras2018progressive}. 

In terms of training, we use  \emph{non-saturating logistic GAN} loss~\cite{karras2020analyzing} with \emph{R1} penalty~\cite{mescheder2018training} to optimize PixelFolder. \emph{Adam} optimizer~\cite{kingma2014adam} is used with a learning rate of $2 \times {10}^{-3}$, and its  hyperparameters $\beta_0$ and $\beta_1$ are set to 0 and 0.99, respectively. The batch size is set to 32 , and the models are trained on 8 NVIDIA V100 32GB GPUs for about four days.

\subsection{Quantitative analysis}
\subsubsection{Comparison with the state-of-the-arts. }
\begin{table}[!t]
\centering
\small
\setlength{\tabcolsep}{3.8mm}{
    \begin{tabular}{llll}
        \toprule
                & \#Parm (M)~$\downarrow$ & GMACs$~\downarrow$ & Speed (im/s)~$\uparrow$ \\
        \midrule
        INR-GAN~\cite{skorokhodov2021adversarial} & 107.03 & 38.76 & \textbf{84.55}  \\
        
        CIPS~\cite{anokhin2021image} & 44.32 & 223.36 & 11.005  \\
        StyleGAN2~\cite{karras2020analyzing} &30.03 & 83.77 & 44.133 \\
        
        \midrule
        PixelFolder (ours) &\textbf{20.84} & \textbf{23.78} & 77.735 \\
       
        \bottomrule
    \end{tabular}
    }
    \caption{Comparison between PixelFolder, StyleGAN2, CIPS and INR-GAN in terms of parameter size (\#Params), computation overhead (GMACs) and inference speed.   Here, ``M'' denotes millions, and ``im/s'' is image per-second. $\uparrow$ denotes that lower is better, while $\downarrow$ is \emph{vice verse}. PixelFolder is much superior than other methods in both model compactness and efficiency, which well validates its innovative designs. }
    \label{tab:efficiency_metric}
\end{table}

\begin{table}[!t]
    \centering
    \setlength{\tabcolsep}{1.3mm}{
    \begin{tabular}{lccc|ccc}
        \toprule
        \multirow{2}{*}{Method} & \multicolumn{3}{c|}{FFHQ, 256$\times$256} & \multicolumn{3}{c}{LSUN Church, 256$\times$256} \\
         & FID~$\downarrow$ &Precision~$\uparrow$ &Recall~$\uparrow$ & FID~$\downarrow$ & Precision~$\uparrow$ & Recall~$\uparrow$ \\
        \midrule
        INR-GAN~\cite{skorokhodov2021adversarial}&4.95& 0.631 &\multicolumn{1}{c|}{0.465} &4.04& 0.590& 0.465\\
        CIPS~\cite{anokhin2021image}&4.38& 0.670 &\multicolumn{1}{c|}{0.407} &2.92& 0.603& 0.474\\
        StyleGAN2~\cite{karras2020analyzing}&3.83~& 0.661 &\multicolumn{1}{c|}{\textbf{0.528}} &3.86 &-&- \\
        \midrule
        PixelFolder(Ours) & \textbf{3.77~} & \textbf{0.683} & \multicolumn{1}{c|}{0.526} & \textbf{2.45~} &\textbf{0.630}&\textbf{0.542}  \\
        \bottomrule
    \end{tabular}
    }
    \caption{The performance comparison of PixelFolder and the SOTA methods on FFHQ \cite{karras2020analyzing} and LSUN Church \cite{radford2015unsupervised}. The proposed PixelFolder not only has better performance than existing pixel synthesis methods, \emph{i.e.}, INR-GAN and CIPS, but also achieves new SOTA performance on both benchmarks. } 
    \label{tab:metrics-results}
\end{table}

We first compare the efficiency of PixelFolder with CIPS \cite{anokhin2021image}, StyleGAN2 \cite{karras2020analyzing} and INR-GAN \cite{skorokhodov2021adversarial} in Tab.~\ref{tab:efficiency_metric}. From this table, we can find that the advantages of PixelFolder in terms of parameter size, computation complexity and inference speed are very obvious. Compared with CIPS, our method can reduce parameters by $53\%$, while the reduction in computation complexity (GMACs) is more distinct, about $89\%$. The inference speed is even improved by about $7\times$. These results strongly confirm the validity of our progressive modeling paradigm and pixel folding operations applied to PixelFolder. Meanwhile, compared with StyleGAN2, the efficiency of PixelFolder is also superior, which reduces $31\%$ parameters and $72\%$ GMACs and speed up the inference by about $76\%$. Also as a multi-stage method, INR-GAN is still inferior to the proposed PixelFolder in terms of parameter size and computation overhead, \emph{i.e.}, nearly $5\times$ more parameters and $1.6\times$ more GMACs. In terms of inference, INR-GAN is a bit faster mainly due to its optimized implementation \footnote{INR-GAN optimizes the CUDA kernels to speed up inference.}. Conclusively, these results greatly confirm the superior efficiency of PixelFolder over the compared image generation methods. 

We further benchmark these methods on FFHQ and LUSN Church, of which results are given in Tab.~\ref{tab:metrics-results}. From this table, we can first observe that on all metrics of two datasets, the proposed PixelFolder greatly outperforms the latest pixel synthesis network, \emph{i.e.}, CIPS\cite{anokhin2021image} and INR-GAN~\cite{skorokhodov2021adversarial}, which strongly validates the motivations of our method about efficient pixel synthesis. Meanwhile, we can observe that compared to StyleGAN2, PixelFolder is also very competitive and obtains new SOTA performance on FFHQ and LSUN Church, \emph{i.e.}, 3.77 FID and 2.45 FID, respectively. Overall, these results suggest that PixelFolder is a method of great potential in image generation, especially considering its high efficiency and low expenditure.

\subsubsection{Ablation studies. }

We further ablates pixel folding operations on FFHQ, of which results are given in Tab.~\ref{tab:ablation}.
Specifically, we replace the pixel folding and unfolding with down-sampling and deconvolution (DeConv.)~\cite{karras2020analyzing}, respectively.

\begin{table}[!t]
        \centering
        \small
        \setlength{\tabcolsep}{1.3mm}{
        \begin{tabular}{lccccccc}
            \toprule
            Settings & \#Parm (M)~$\downarrow$ & GMACs~$\downarrow$& FID~$\downarrow$&Precision~$\uparrow$ &Recall~$\uparrow$ \\
            \midrule
            \textbf{Fold+Unfold (base)}& \textbf{20.84} & \textbf{23.78} & \textbf{5.49} & \textbf{0.679} & \textbf{0.514}\\
            Fold+DeConv & 29.41 & 86.53 & 5.60 & 0.667 & 0.371  \\
            Down-Sampling+DeConv & 29.21 & 89.38 & 5.53 & \ 0.679 & 0.456  \\
            
            \bottomrule
        \end{tabular}
        }
        \caption{Ablation study on FFHQ. The models of all settings are trained with 200k steps for a quick comparison. These results show the obvious advantages of pixel folding (Fold+Unfold) over down-sampling and DeConv. }
        \label{tab:ablation}
    \end{table}

\begin{table}[!t]
        \centering
        \small
        \setlength{\tabcolsep}{1.mm}{
        \begin{tabular}{lccccccc}
            \toprule
            Settings & \#Parm (M)~$\downarrow$ & GMACs~$\downarrow$& FID~$\downarrow$&Precision~$\uparrow$ &Recall~$\uparrow$ \\
            \midrule
            \textbf{PixelFolder} & 20.84 & 23.78 & \textbf{4.78} & \textbf{0.602} & \textbf{0.517} \\
            $w/o$ coordinate embeddings & \textbf{20.32} & \textbf{23.64} & 4.95 & 0.598 & 0.500\\
            $w/o$ multi-stage connection & 20.84 & 23.78 & 5.46 & 0.532 & 0.441 \\
            
            \bottomrule
        \end{tabular}
        }
        \caption{Ablation study on LSUN Church. The models of all settings are trained with 200k steps for a quick comparison. “\emph{$w/o$ design}” is not cumulative and only represents the performance of PixelFolder without this design/setting. }
        \label{tab:ablation-self}
    \end{table}

From these results, we can observe that although these operations can also serve to reduce or recover tensor resolutions, their practical effectiveness is much inferior than our pixel folding operations, \emph{e.g.} 5.49 FID (fold+unfold) \emph{v.s.} 8.36 FID (down-sampling+DeConv). These results greatly confirm the merit of pixel folding in preserving pixel-wise independence, which can help the model exploit pixel-wise prior knowledge. In Tab.~\ref{tab:ablation-self}, we examine the initialization of pixel tensor and the impact of multi-stage connection. From this table, we can see that only using Fourier features without coordinate embeddings slightly reduces model performance, but this impact is smaller than that in CIPS~\cite{anokhin2021image}. This result also subsequently suggests that PixelFolder do not rely on large parameterized tensors to store pixel-wise prior knowledge, leading to better model compactness.
Meanwhile, we also notice that without the multi-stage connection, the performance drops  significantly, suggesting the importance of joint multi-scale pixel regression, as discussed in Sec.~\ref{subsec:overview}. 
Overall, these ablation results well confirm the effectiveness of the designs of PixelFolder.

\subsection{Qualitative analysis}
To obtain deep insight into the proposed PixelFolder, we further visualize its synthesized images as well as the ones of other SOTA methods.

\begin{figure}[!t]
    \centering
    \includegraphics[width=0.98\linewidth]{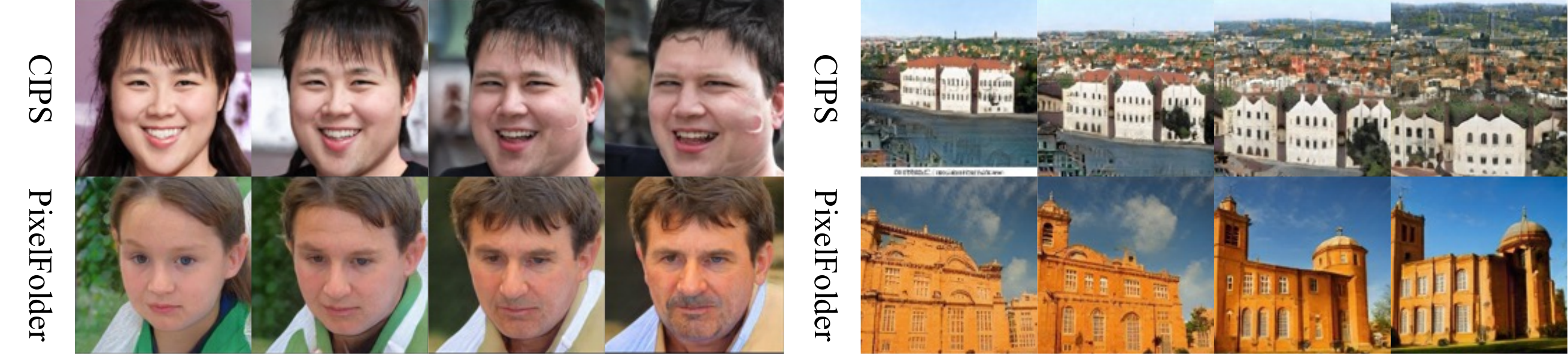}
    \caption{
    Comparison of the image interpolations by CIPS \cite{anokhin2021image} and PixelFolder. The interpolation is computed by $\mathbf{z}=\alpha \mathbf{z}_1+(1-\alpha)\mathbf{z}_2$, where $\mathbf{z}_1$ and $\mathbf{z}_2$ refer to the left-most and right-most samples, respectively. 
    }
    \label{fig:interp}
\end{figure}

\begin{figure}[!htbp]
    \centering
    \subcaptionbox{
    FFHQ-eyeglasses \label{subfig:eg}}{
    \includegraphics[width =.475\linewidth]{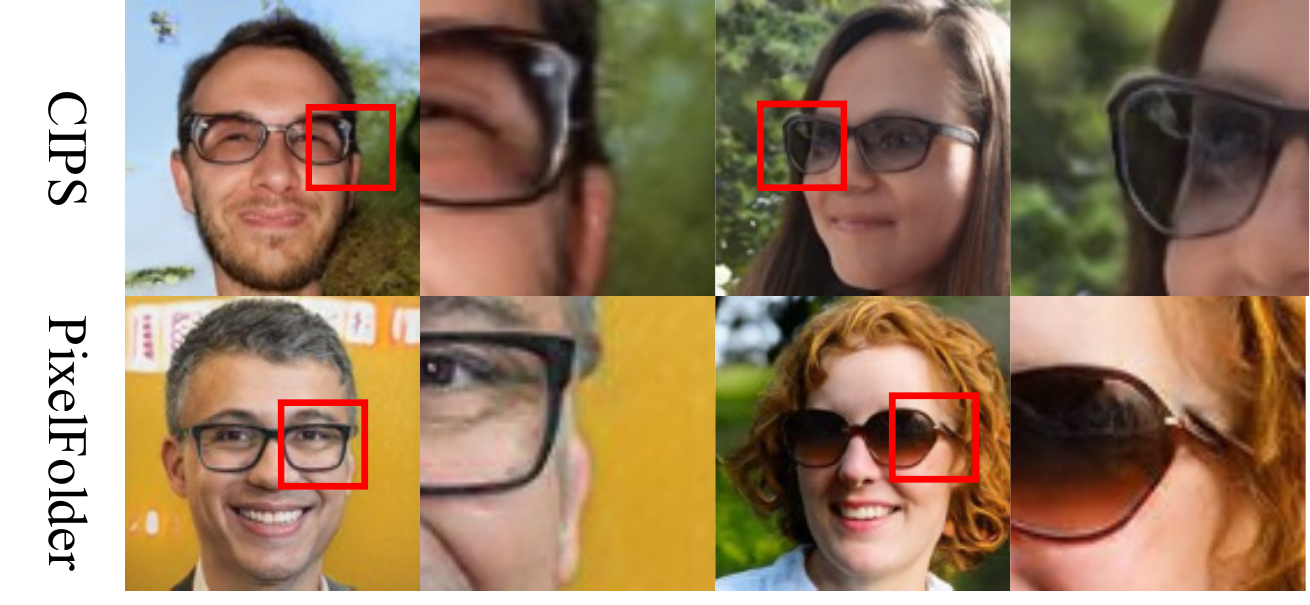}
    }
    \hfill
    \subcaptionbox{FFHQ-headwear \label{subfig:hw}}{
    \includegraphics[width = .475\linewidth]{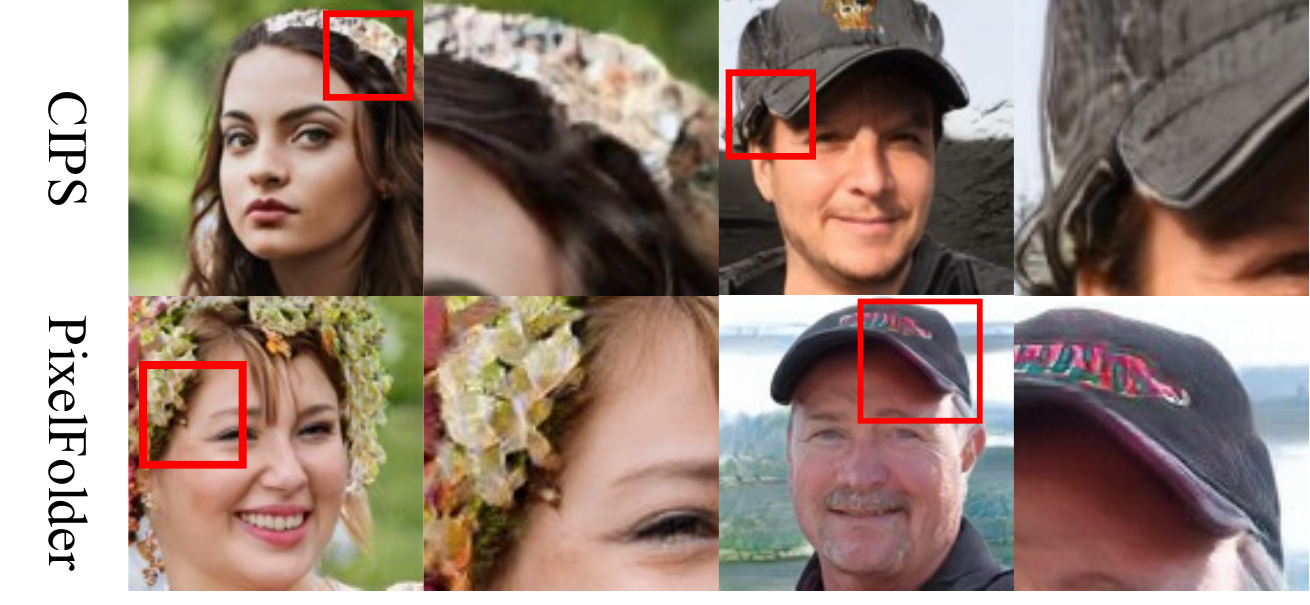}
    }
    \hfill
    \subcaptionbox{LSUN Church-wavy texture \label{subfig:wt}}{
    \includegraphics[width = .475\linewidth]{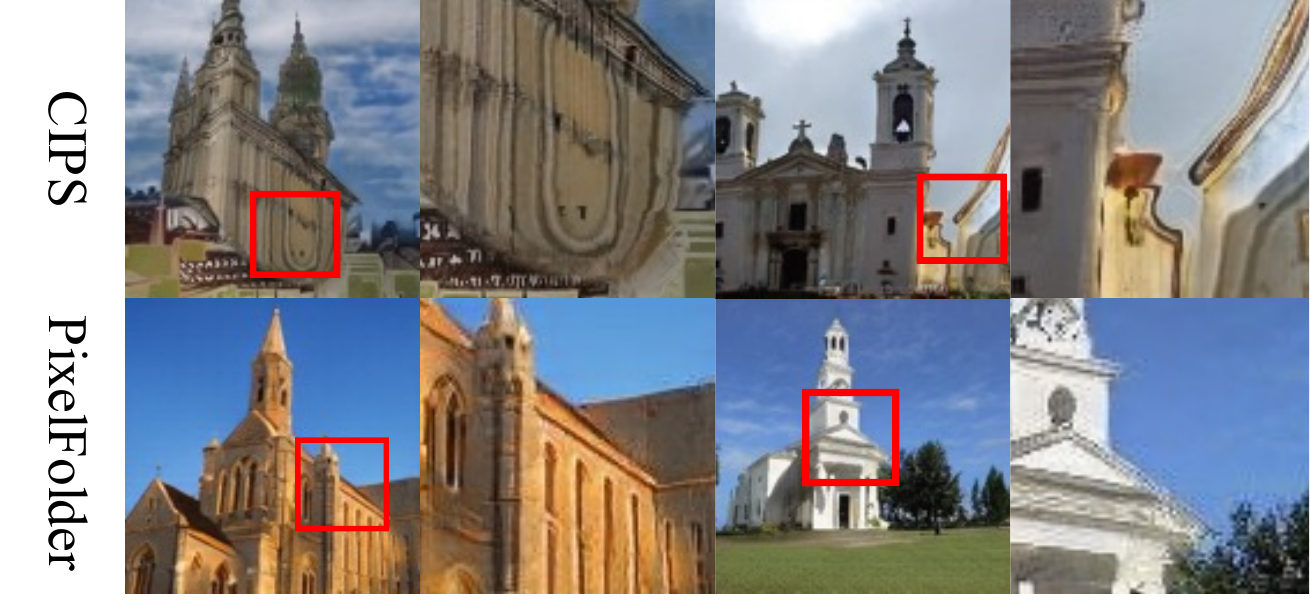}
    }
    \hfill
    \subcaptionbox{LSUN Church-pixel offset \label{subfig:po}}{
    \includegraphics[width = .475\linewidth]{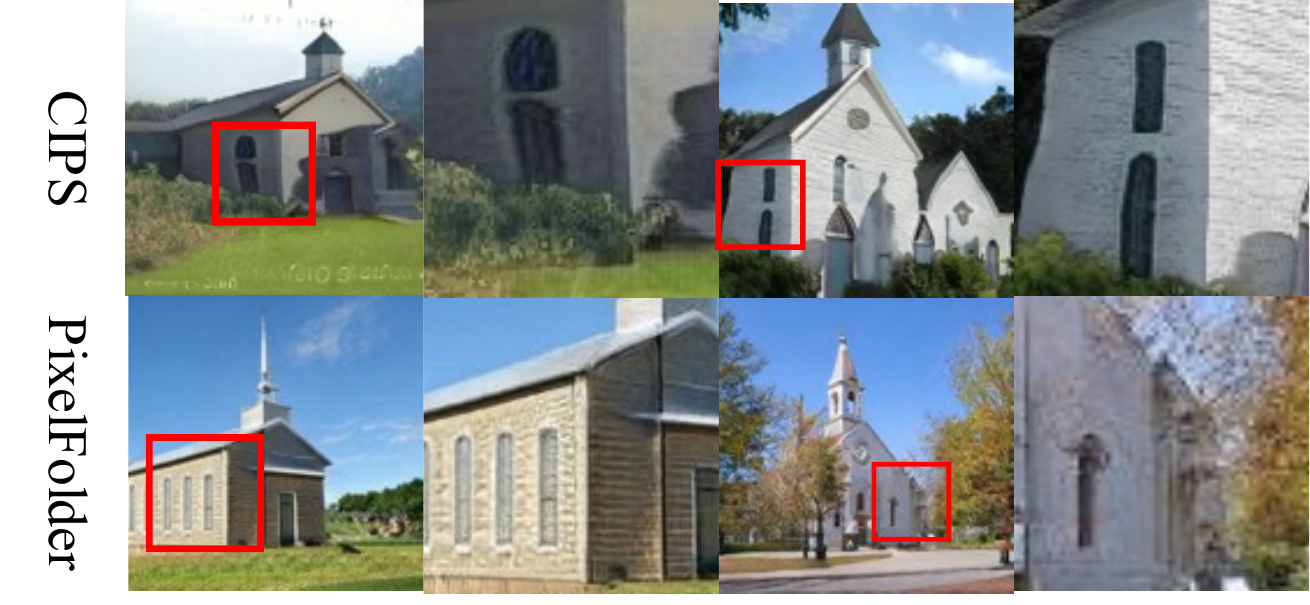}
    }
    \caption{
    Comparison of the generated images by CIPS~\cite{anokhin2021image} and PixelFolder on FFHQ and LSUN Church. The overall quality of images generated by PixelFolder is better than that of CIPS. Meanwhile, PixelFolder can better handle the local imagery incongruity, confirming the effectiveness of its designs. 
    }
    \label{fig:cips-pixelfolder}
\end{figure}

\subsubsection{Comparison with CIPS. }
We first compare the image interpolations of PixelFolder and CIPS on two benchmarks, \emph{i.e.}, FFHQ and LSUN Church, as shown in Fig.~\ref{fig:interp}. It can be obviously seen that the interpolations by PixelFolder are more natural and reasonable than those of CIPS, especially in terms of local imaging. 
We further present more images synthesized by two methods in Fig.~\ref{fig:local_incongruities} and Fig. ~\ref{fig:cips-pixelfolder}. 
From these examples, a quick observation is that the overall image quality of PixelFolder is better than CIPS. 
The synthesized faces by PixelFolder look more natural and vivid, which also avoid obvious deformations. Meanwhile, the surroundings and backgrounds of the generated church images by PixelFolder are more realistic and reasonable, as shown in Fig.~\ref{subfig:wt}-\ref{subfig:po}. 
In terms of local imaging, the merit of PixelFolder becomes more obvious. As discussed in this paper, CIPS is easy to produce local pixel incongruities due to its relatively independent pixel modeling strategy \cite{anokhin2021image}. 
This problem is reflected in its face generations, especially the hair details. 
In contrast, PixelFolder well excels in local imaging, such as the synthesis of accessories and hat details, as shown in Fig.~\ref{subfig:eg}-\ref{subfig:hw}. Meanwhile, CIPS is also prone to wavy textures and distortions in the church images, while these issues are greatly  alleviated by PixelFolder. 
Conclusively, these findings well validate the motivations of PixelFolder for image generation. 

\begin{figure*}[!t]
    \centering
    \includegraphics[width=0.8\linewidth]{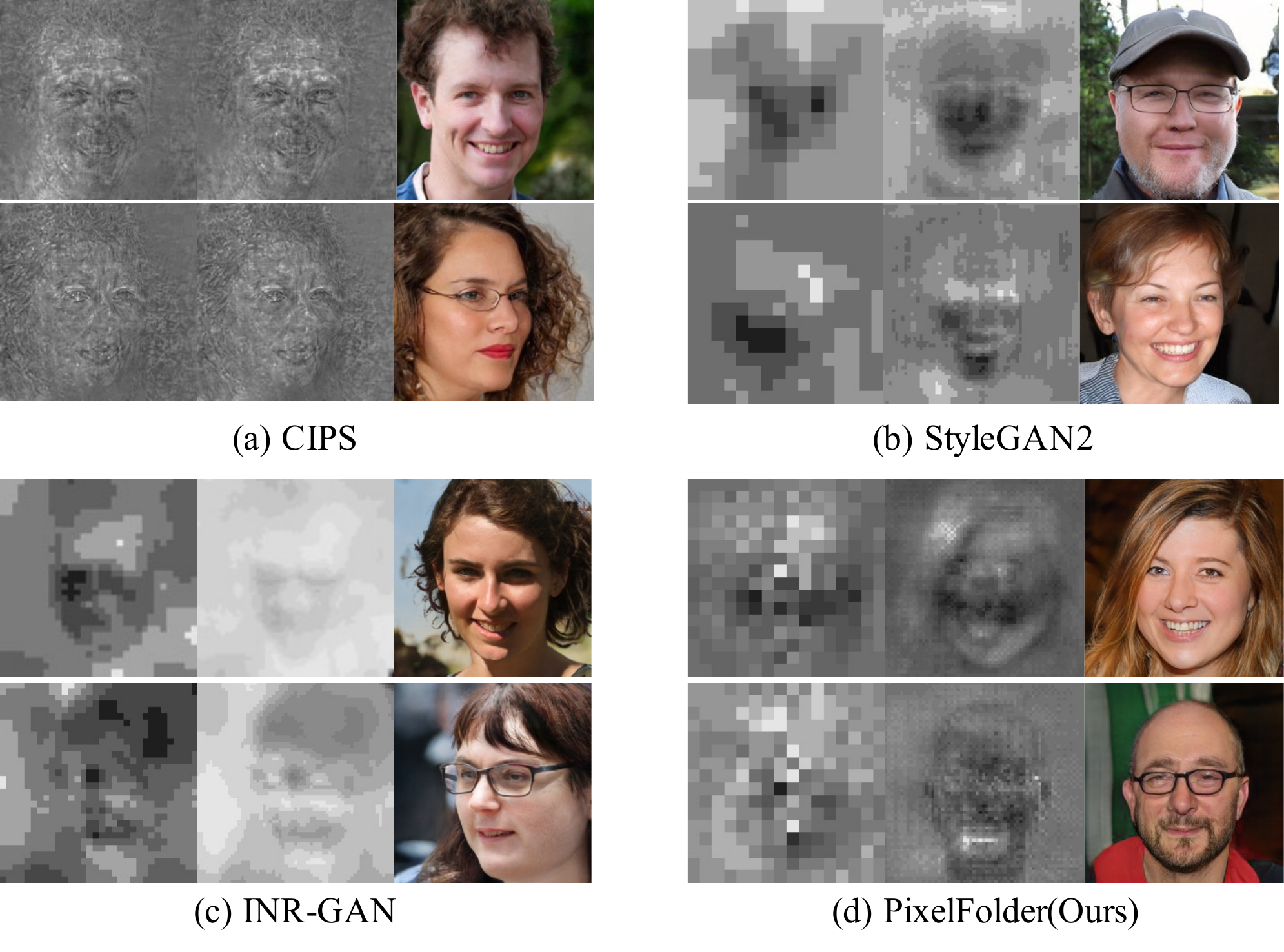}
    \caption{Comparison of the stage-wise synthesis by the SOTA methods and PixelFolder. The color spaces of the first two hidden images are uniformly adjusted for better observation. 
    We chose the hidden images of all methods from the same number of convolution layers. Pixel-synthesis based methods, such as CIPS~\cite{anokhin2021image} and PixelFolder, present more interpretable results in initial steps, where PixelFolder can also provide better outline details.}
    \label{fig:visualize-every-block}
\end{figure*}

\subsubsection{Comparison of stage-wise visualizations. }
We also compare PixelFolder with CIPS, StyleGAN2 and INR-GAN by visualizing their stage-wise results, as shown in Fig.~\ref{fig:visualize-every-block}. From these examples, we can first observe that the intermediate results of other progressive methods, \emph{i.e.}, StyleGAN2 and INR-GAN, are too blurry to recognize. In contrast, PixelFolder and CIPS can depict the outline of generated faces in the initial and intermediate stages. 
This case suggests that PixelFolder and CIPS can well exploit the high-frequency information provided by Fourier features~\cite{anokhin2021image}, verifying the merits of end-to-end pixel regression. 
We can also see that PixelFolder can learn more details than CIPS in the intermediate features, which also suggests the superior efficiency of PixelFolder in face generation. Meanwhile, the progressive refinement (from left to right) also makes PixelFolder more efficient than CIPS in computation overhead and memory footprint.  We attribute these advantages to the pixel folding operations and the multi-stage paradigm of PixelFolder, which can help the model exploit prior knowledge in different generation stages.

\begin{figure*}[!t]
    \centering
    \subcaptionbox{folding+unfolding\label{subfig:pxf-base}}{
    \includegraphics[width =.3\linewidth]{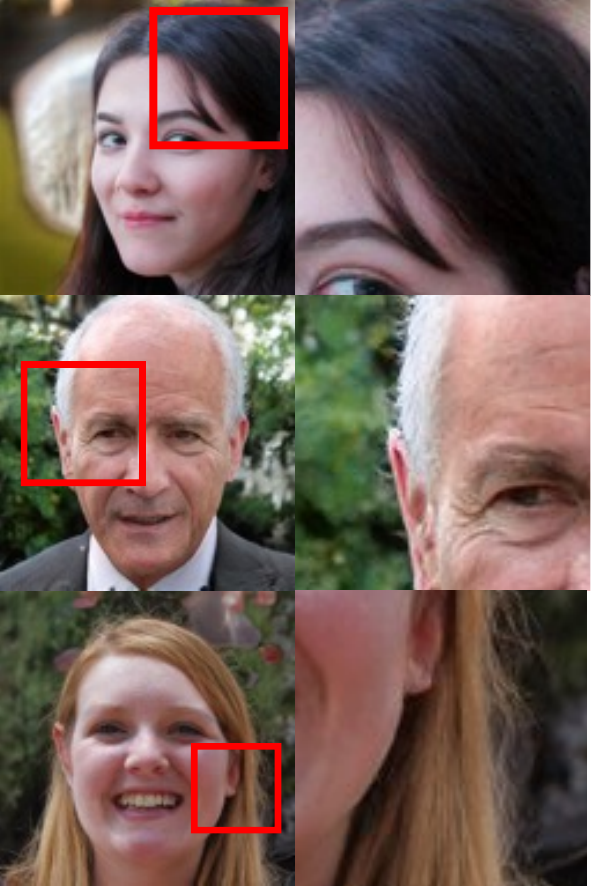}
    }
    \hfill
    \subcaptionbox{folding+DeConv\label{subfig:pxf_fdc}}{
    \includegraphics[width = .3\linewidth]{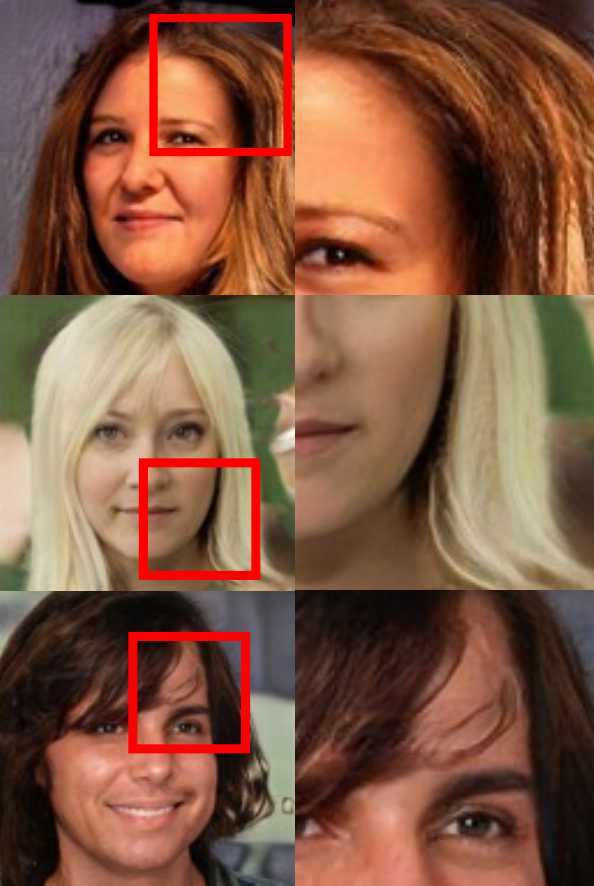}
    }
    \hfill
    \subcaptionbox{downsample+DeConv\label{subfig:pxf_dsdc}}{
    \includegraphics[width = .3\linewidth]{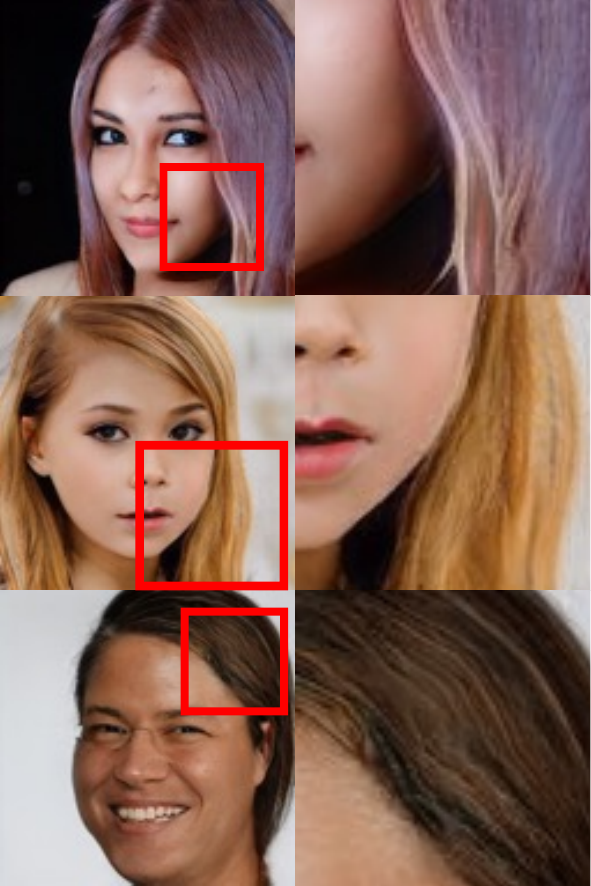}
    }
    \caption{Comparisons of PixelFolder with pixel folding operations (folding+unfolding) and the alternatives (\emph{i.e.}, folding+DeConv. and down-sampling+DeConv). Compared with these alternatives, pixel folding operations can well preserve pixel-wise prior knowledge for generation, leading to much better image quality. Meanwhile, pixel folding can also well tackle with local imagery incongruities. }
    \label{fig:v_abl}
\end{figure*}

\subsubsection{Comparison of pixel folding and its alternatives.  }
In Fig.~\ref{fig:v_abl}, we visualize the generations of PixelFolder with pixel folding operations and the alternatives mentioned in Tab.~\ref{tab:ablation}. From these examples, we can find that although down-sampling and DeConv. can also serve to change the resolution of hidden pixel tensors, their practical effectiveness is still much inferior than that of pixel folding. 
We attribute these results to the unique property of pixel folding in preserving pixel-wise prior knowledge for end-to-end pixel regression. 
Meanwhile, we also note that when using these alternatives, there is still the problem of local image incongruity, which however can be largely avoided by pixel foldings. 
These results greatly validate the motivation and effectiveness of the pixel folding operations.

\section{Conclusions}
In this paper, we propose a novel pixel synthesis network towards efficient image generation, termed \emph{PixelFolder}. 
Specifically, PixelFolder considers the pixel synthesis as a problem of progressive pixel regression, which can greatly reduce the excessive overhead caused by large tensor transformations. Meanwhile, we also apply novel \emph{pixel folding} operations to further improve model efficiency while preserving the property of end-to-end pixel regression. 
With these novel designs, PixelFolder requires much less computational and memory overhead than the latest pixel synthesis methods, such as CIPS and INR-GAN. 
Meanwhile, compared with the state-of-the-art method StyleGAN2, PixelFolder is also more efficient. 
With much higher efficiency, the proposed PixelFolder exhibits new SOTA performance on FFHQ and LSUN Church benchmarks, \emph{i.e.}, 3.77 FID and 2.45 FID, respectively, yielding a great potential in image generation.  \\

\subsubsection{Acknowledgements}
This work is supported by the National Science Fund for Distinguished Young (No.62025603), the National Natural Science Foundation of China (No.62025603, No. U1705262, No. 62072386, No. 62072387, No. 62072389, No. 62002305,  No.61772443, No. 61802324 and No. 61702136) and Guangdong Basic and Applied Basic Research Foundation (No.2019B1515120049).

\title{PixelFolder: An Efficient Progressive Pixel Synthesis Network for Image Generation}
\author{}
\institute{}
\subtitle{Supplementary Materials}
\maketitle
\appendix
\renewcommand\thetable{\Alph{table}}
\renewcommand\thefigure{\Alph{figure}}
\section{More Experiments}
\subsection{Comparison of additional metrics for image quality}
To further validate our PixelFolder in terms of image quality, we report two new indicators, namely Inception Score (IS)
and Perceptual Path Length (PPL). The IS was shown to correlate well with the human scoring for judging the realism of generated images on CIFAR-10 dataset~\cite{salimans2016improved}, the higher IS score indicators the better image quality. The PPL indicates the smoothness of the mapping from a latent space to the output images, which is very important for high quality image generation~\cite{karras2019style}. The smooth mapping will have a small PPL. 
In Tab. \ref{tab:metrics-is-ppl}, the overall performance of PixelFolder is still better. 

\begin{table}[!b]\small
    \centering
    \setlength{\tabcolsep}{5.5mm}{
    \begin{tabular}{lcc|cc}
        \toprule
        \multirow{2}{*}{Method} & \multicolumn{2}{c|}{FFHQ, 256$\times$ 256} & \multicolumn{2}{c}{LSUN Church, 256 $\times$ 256} \\
         & IS~$\uparrow$ &PPL~$\downarrow$ & IS~$\uparrow$ & PPL~$\downarrow$ \\
        \midrule
        INR-GAN & 4.51$\pm$0.08 & \textbf{62} & 2.69$\pm$0.02 & 529 \\
        CIPS & 4.75$\pm$0.06 & 273 & 2.73$\pm$0.02 & 3132 \\
        StyleGAN2 & 4.86$\pm$0.04 & 209 & 2.78$\pm$0.02 & 350 \\
        \midrule
        PixelFolder & \textbf{5.04$\pm$0.07} & 197 & \textbf{2.79$\pm$0.03} & \textbf{307} \\
        
        \bottomrule
    \end{tabular}\vspace{2mm}
    }
    \caption{The comparison of inception score (IS) and perceptual path length (PPL) on FFHQ and LSUN Church. The higher IS score and the lower PPL score indicate the better performance in terms of image quality. These results further validate the superior of the proposed PixelFolder. } 
    \label{tab:metrics-is-ppl}
    
\end{table}

\begin{table}[!t]\small
    \centering
    \setlength{\tabcolsep}{5.8mm}{

    \begin{tabular}{lcccc}
        \toprule
        \multirow{2}{*}{Method} & \multicolumn{2}{c}{FFHQ, $512\times 512$} & \multicolumn{2}{c}{FFHQ, $1024\times 1024$}\\
         & $\#$Params$\downarrow$ & FID $\downarrow$&$\#$Params$\downarrow$ & FID$\downarrow$ \\
        \midrule
        INR-GAN & 112.43 & - & 117.30 & 16.32  \\
        CIPS & 145.12 & - & 574.32 & -  \\
        StyleGAN2 & 30.28 & 4.15 & 30.37 & \textbf{2.84}  \\
        \midrule
        PixelFolder & \textbf{20.84} & \textbf{4.08} & \textbf{21.14} & 2.98 \\
       
        \bottomrule
    \end{tabular}\vspace{2mm}
    }
    \caption{Comparison of high-resolution image generation. The proposed PixelFolder has better performance with fewer parameters than other methods. } 
    \label{tab:metrics-res}
    
\end{table}

\subsection{Comparison of high-resolution image generation}
In Tab. \ref{tab:metrics-res}, we compare the performance of high-resolution image generation between the proposed PixelFolder and other SOTA methods, \emph{i.e.} $512\times 512$ and $1024\times 1024$. From these results, we observe that PixelFolder still has competitive performance in terms of model efficiency and image quality. Compared to StyleGAN2, PixelFolder obtains competitive FID values with only two-thirds of its parameters, and especially at $512\times 512$ resolutions where PixelFolder obtains better performance. Note that the proposed PixelFolder is directly
implemented without any tricks like adding additional noises. INR-GAN and CIPS are out of GPU memory to train. 

\begin{minipage}{\textwidth}
\hspace{-8mm}
\begin{minipage}[!t]{0.5\textwidth}
\makeatletter\def\@captype{table}
\centering
\setlength{\tabcolsep}{3.2mm}{
\begin{tabular}{lcc}
    \toprule
    Method & Cat & Bedroom \\
    \midrule
    INR-GAN & - & 3.40 \\
    CIPS & 12.86 & - \\
    StyleGAN2 & 9.14 & \textbf{2.65} \\
    \midrule
    PixelFolder & \textbf{8.41} & 3.71 \\
    \bottomrule
\end{tabular}}
\caption{Comparison of FID scores on LSUN Cat and LSUN Bedroom. }
\label{tab:metrics-more}
\end{minipage}
\begin{minipage}[!t]{0.48\textwidth}\hspace{8mm}
\makeatletter\def\@captype{table}
\centering
\setlength{\tabcolsep}{1.0mm}{
\begin{tabular}{lccc}
    \toprule
    Method & FFHQ & Church  \\
    \midrule
    UDM(RVE)+ST [a] & 5.54 & -  \\
    UnleashTR [b] & 6.11 & 4.07  \\
    \midrule
    StyleGAN2 & 3.83 & 3.86  \\
    PixelFolder & \textbf{3.77} & \textbf{2.46}  \\
    
    \bottomrule
\end{tabular}
}
\caption{Comparison with diffusion models in terms of FID score. }
\label{tab:metrics-diffusion}
\end{minipage}
\end{minipage}

\subsection{Comparison on more datasets}
We conduct additional experiments on two popular datasets as reported in Tab. \ref{tab:metrics-more}, \emph{i.e.} LSUN cat and LSUN bedroom. The proposed PixelFolder still obtains competitive performance, which further demonstrates the generalizability of our PixelFolder.

\subsection{Comparison with more typical methods} 
In addition to some GAN-based approaches~\cite{anokhin2021image,karras2020analyzing}, we further compare PixelFolder with the recently  popular diffusion models~\cite{kim2021exploiting,bond2021unleashing} in Tab. \ref{tab:metrics-diffusion}, where the merits of our method still can be seen.

\section{Experiment details}
To ablate the pixel folding operations, we compare them with the alternatives of \emph{down-sampling} and \emph{deconvolution} (DeConv) in Tab.3 of the main paper. 
We can see that both alternative models, \emph{i.e.}, Fold+DeConv and Down-Sampling+DeConv,  have more parameters than Fold+Unfold (base). The main reason for the parameter gap is that the pixel unfolding operation reduces the number of channels of the hidden tensors, thus the corresponding convolution kernel has fewer parameters (see \textcolor{blue}{blue} font in Tab.~\ref{tab:tensor_shape}).

\begin{table}[!t]
    \centering
    \small
    \setlength{\tabcolsep}{.8mm}{
    \renewcommand{\arraystretch}{1.8}
    \begin{tabular}{l|c|c|c}
    \toprule
     Layers & Fold+UnFold (base) & Fold+DeConv & DownS.+DeConv \\
    \hline
    Initialization & $16\times16\times512$  & $16\times16\times512$ & $16\times16\times512$ \\
    \hline
    projection & $16\times16\times32$ & $16\times16\times32$ & - \\
    \hline
    Folding$\times2$/DownS. & $4\times4\times512$ & $4\times4\times512$ & $4\times4\times512$ \\
    \hline
    ModConv\mypm{0}/DeConv\mypm{0} & \makecell[c]{$4\times4\times512$\\($3\times3\times512\times512$)} & \makecell[c]{\mypmRED{$8\times8\times512$}\\($3\times3\times512\times512$)} & \makecell[c]{\mypmRED{$8\times8\times512$}\\($3\times3\times512\times512$)} \\
    \hline
    Unfolding\mypm{0} & \mypmRED{$8\times8\times128$} & - & - \\
    \hline
    ModConv\mypm{1} & \makecell[c]{$8\times8\times512$\\\mypmBLUE{($3\times3\times128\times512$)}} & \makecell[c]{$8\times8\times512$\\\mypmBLUE{($3\times3\times512\times512$)}} & \makecell[c]{$8\times8\times512$\\\mypmBLUE{($3\times3\times512\times512$)}} \\
    \hline
    ModConv\mypm{2}/DeConv\mypm{1} & \makecell[c]{$8\times8\times512$\\($3\times3\times512\times512$)} & \makecell[c]{\mypmRED{$16\times16\times512$}\\($3\times3\times512\times512$)} & \makecell[c]{\mypmRED{$16\times16\times512$}\\($3\times3\times512\times512$)} \\
    \hline
    Unfolding\mypm{1} & \mypmRED{$16\times16\times128$} & - & - \\
    \hline
    ModConv\mypm{3} & \makecell[c]{$16\times16\times512$\\\mypmBLUE{($3\times3\times128\times512$)}} & \makecell[c]{$16\times16\times512$\\\mypmBLUE{($3\times3\times512\times512$)}} & \makecell[c]{$16\times16\times512$\\\mypmBLUE{($3\times3\times512\times512$)}} \\
    \hline
    ToRGB & \makecell[c]{$16\times16\times3$\\($1\times1\times3\times512$)} & \makecell[c]{$16\times16\times3$\\($1\times1\times3\times512$)} & \makecell[c]{$16\times16\times3$\\($1\times1\times3\times512$)} \\

    \bottomrule
    \end{tabular}
}
    \vspace{2mm}
    \caption{The tensor shapes of the output of layers in the first row. ($k\times k\times c_{in} \times c_{out}$) represents the kernel size of corresponding convolution layer. The \textcolor{red}{red} font indicates the tensor shape before feeding into ModConv or DeConv, and the \textcolor{blue}{blue} font indicates the kernel size of ModConv or DeConv with red tensor as input. }\vspace{-7mm}
    \label{tab:tensor_shape}
\end{table}

\begin{table}[!t]
        \centering
        \small
        \setlength{\tabcolsep}{1.mm}{
        \begin{tabular}{lccccccc}
            \toprule
            Settings & \#Parm (M)~$\downarrow$ & GMACs~$\downarrow$& FID~$\downarrow$&Precision~$\uparrow$ &Recall~$\uparrow$ \\
            \midrule
            \textbf{Fold+Unfold (base)}& \textbf{20.84} & \textbf{23.78} & \textbf{5.49} & \textbf{0.679} & \textbf{0.514}\\
            \midrule
            Fold+DeConv & 29.41 & 86.53 & 5.60 & 0.667 & 0.371  \\
            Fold+DeConv-sc & 12.66 & 12.12 & 8.41 & 0.643 & 0.366  \\
            \midrule
            Down-Sampling+DeConv & 29.21 & 89.38 & 5.53 & \ 0.679 & 0.456  \\
            Down-Sampling+DeConv-sc & 12.46 & 12.97 & 8.36 & \ 0.670 & 0.442  \\
            
            \bottomrule
        \end{tabular}
        }
        \vspace{2mm}
        \caption{Additional ablation study on FFHQ. The models of all settings are trained with 200k steps for a quick comparison. "sc" is short for "shape-consistent". The results further prove the advantages of pixel folding operations. }
        \vspace{-8.3mm}
        \label{tab:more_ablation}
    \end{table}

In order to keep the shape of the hidden tensor with \textcolor{red}{red} color consistent, we further modify DeConv\mypm{0} and DeConv\mypm{1}, \emph{i.e.}, the kernels are replaced with $3\times3\times512\times128$ and $3\times3\times128\times512$, respectively. The results of the shape-consistent alternatives are shown in Tab.~\ref{tab:more_ablation}. Although Fold+DeConv-sc and Down-Sampling+DeConv-sc guarantee the consistency of the shape of pixel tensors, their performance lags far behind that of Fold+Unfold (base), which further demonstrates the effectiveness of pixel folding operations.

\section{Additional samples}
\begin{figure}[!h]
    \centering
    \centering
    \subcaptionbox{
    FFHQ. \label{subfig:iffhq}}{
    \vspace{-1mm}
    \includegraphics[width =.8\linewidth]{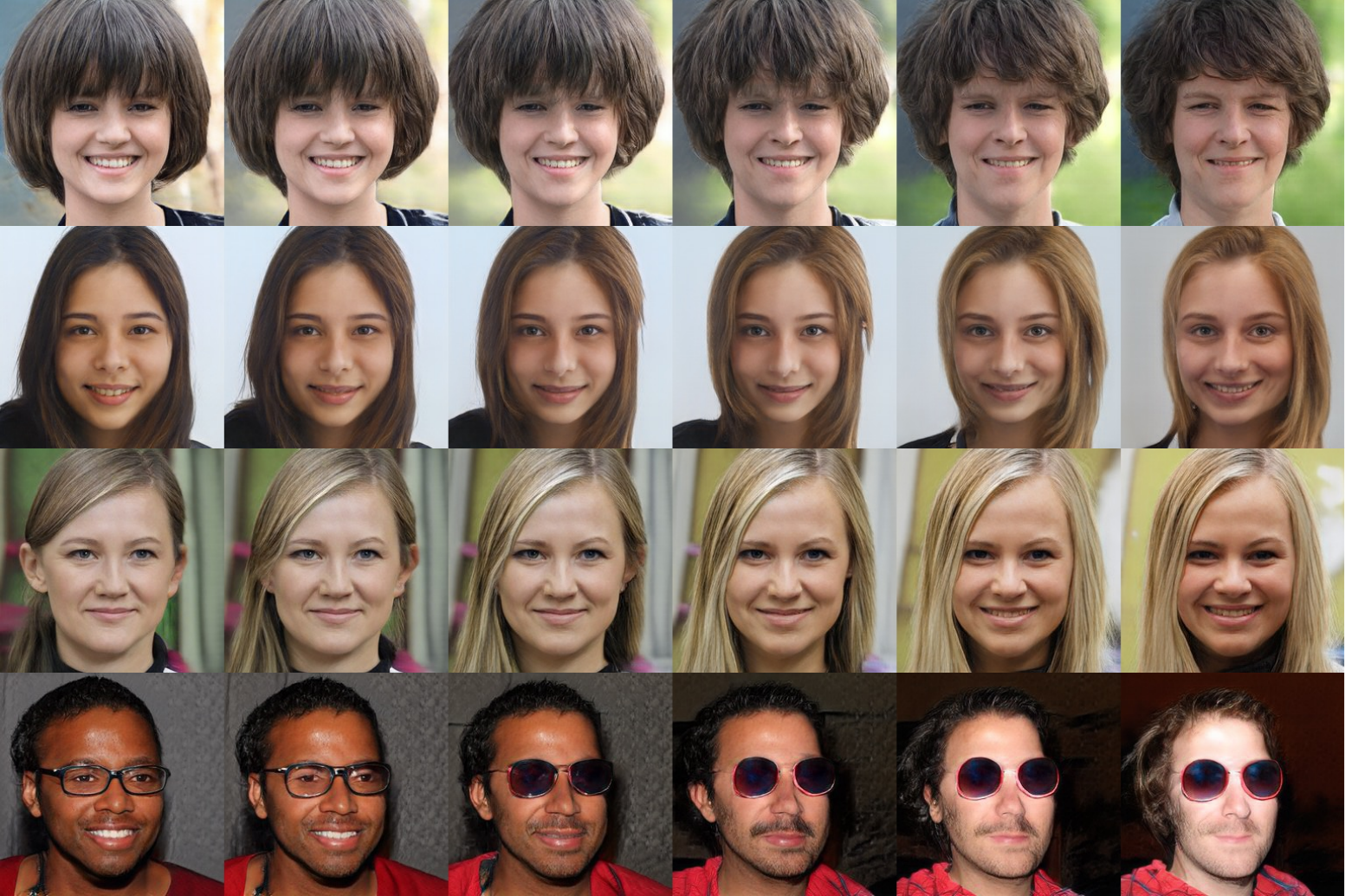}
    }
    \vfill
    \subcaptionbox{LSUN Church. \label{subfig:ichurch}}{
    \vspace{-1mm}
    \includegraphics[width = .8\linewidth]{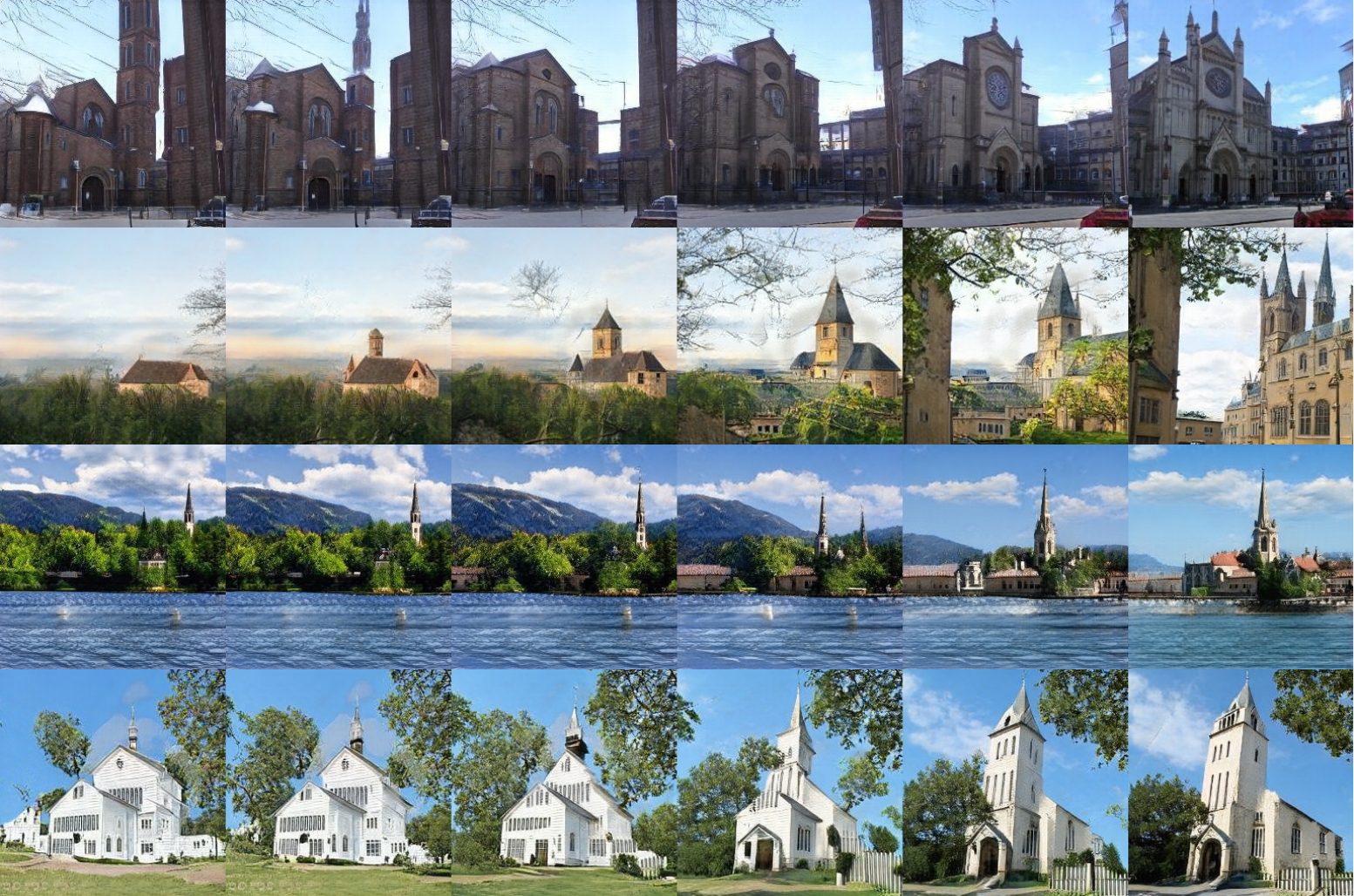}
    }
    
    \vspace{-2mm}
    \caption{
    Interpolations by PixelFolder on FFHQ and LSUN Church. The factor $\alpha$ of the interpolations is set to smaller than that of Fig.4 of the main paper. These interpolations further demonstrate the generalization capability of PixelFolder when the input noise fluctuates slightly. 
    }
    \vspace{-0.6cm}
    \label{fig:more_inter}
\end{figure}

\begin{figure}[!htbp]
    \centering
    \subcaptionbox{
    FFHQ. \label{subfig:ffhq}}{
    \vspace{-1mm}
    \includegraphics[width =.475\linewidth]{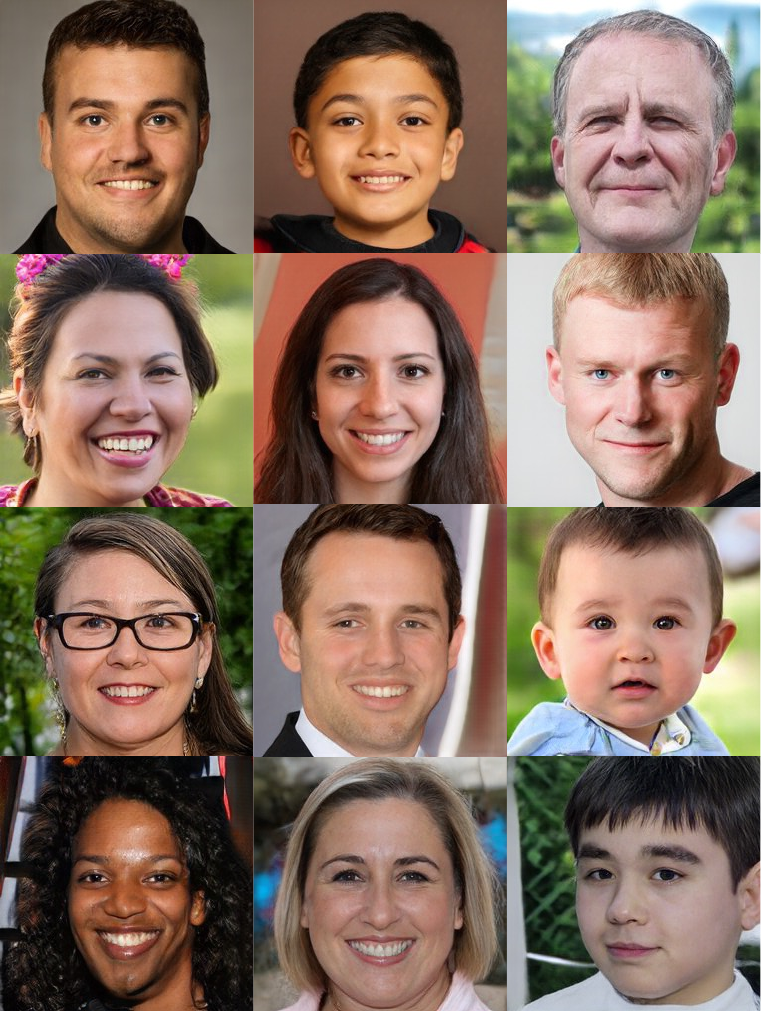}
    }
    \hfill
    \subcaptionbox{LSUN Church. \label{subfig:church}}{
    \vspace{-1mm}
    \includegraphics[width = .475\linewidth]{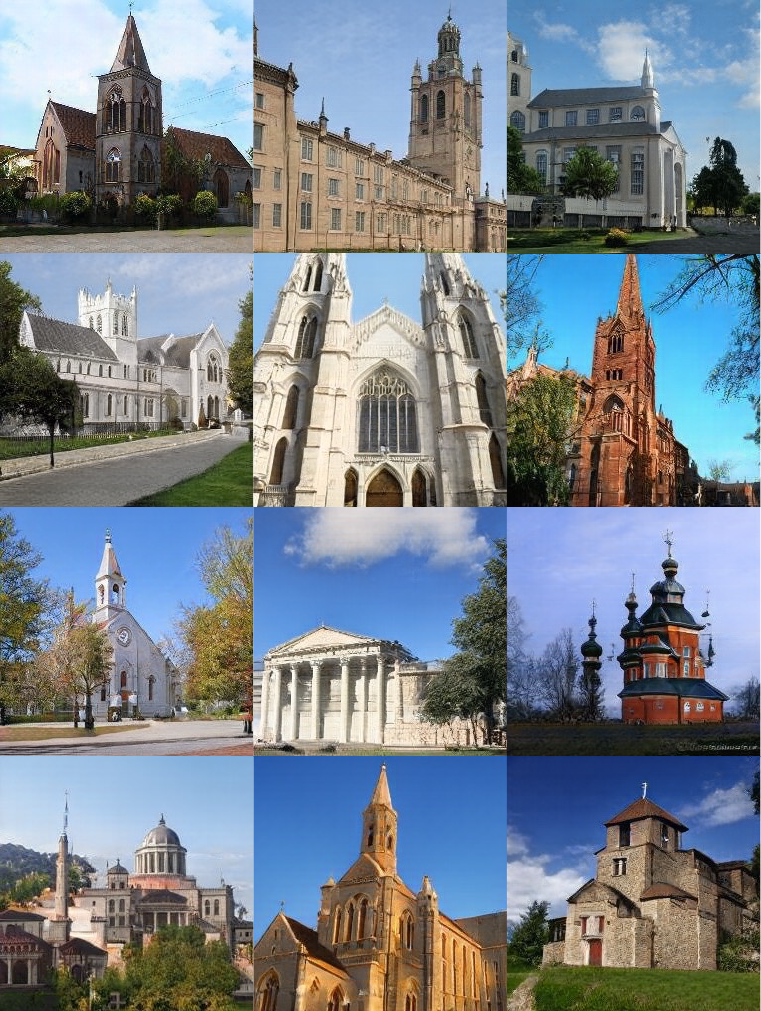}
    }
    \hfill
    \subcaptionbox{LSUN Bedroom. \label{subfig:bedroom}}{
    \vspace{-1mm}
    \includegraphics[width = .475\linewidth]{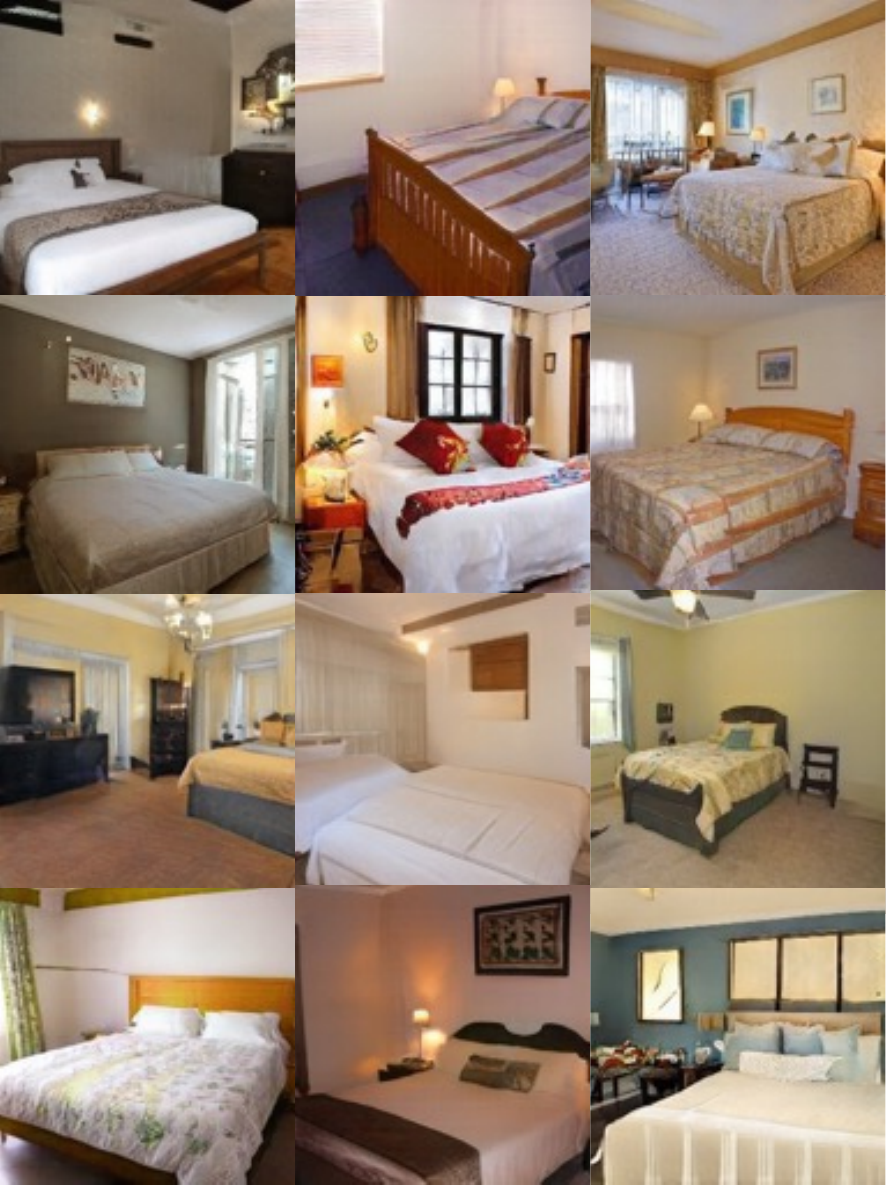}
    }
    \hfill
    \subcaptionbox{LSUN Cat. \label{subfig:cat}}{
    \vspace{-1mm}
    \includegraphics[width = .475\linewidth]{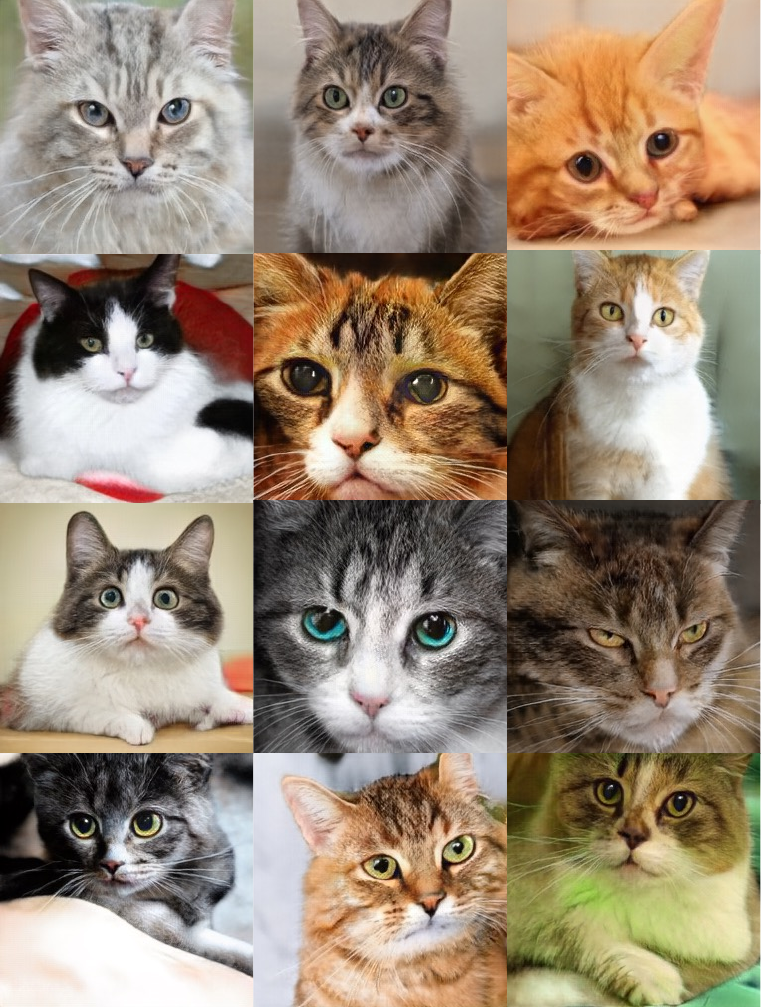}
    }
    \vspace{-2mm}
    \caption{
    Additional samples of PixelFolder on different datasets, \emph{i.e.}, FFHQ, LSUN Church, LSUN Bedroom and LSUN Cat. 
    }
    \vspace{-0.6cm}
    \label{fig:more_samples}
\end{figure}

\begin{figure}
    \centering
    \includegraphics[width = .99\linewidth]{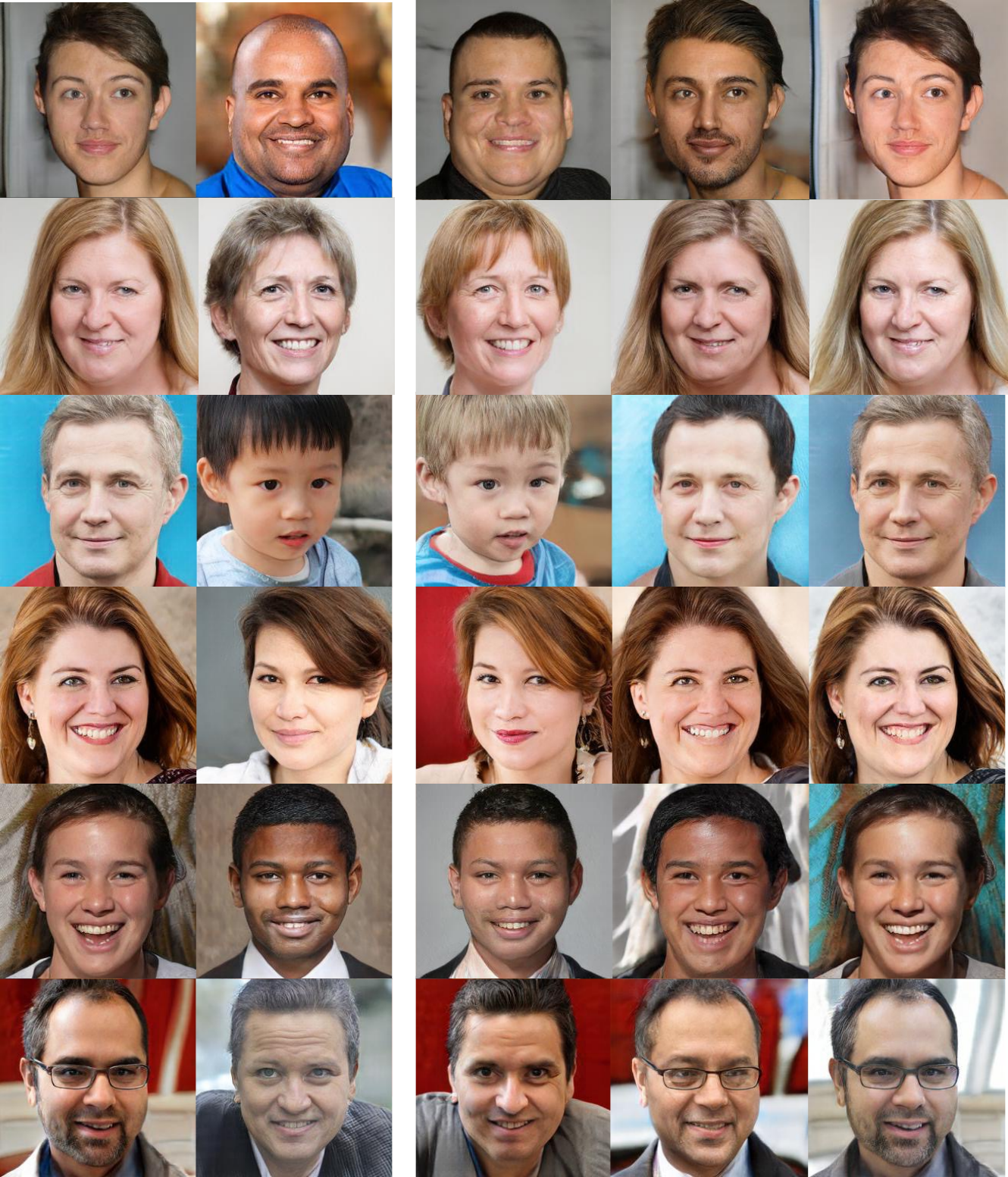}
    \caption{Stage-wise style mixing of our PixelFolder trained on FFHQ. The left two columns are the images generated by latent code $z_1$ and $z_2$, respectively. The three columns on the right are the images generated by replacing $z_1$ at stage1, stage2 and stage3 with the corresponding latent code $z_2$.}
    \label{fig:style_mixing}
\end{figure}

In Fig.~\ref{fig:more_inter}, we provide more interpolations on FFHQ and LSUN Church to further demonstrate the generalization capability of PixelFolder. We also provide additional samples on different datasets in Fig.~\ref{fig:more_samples}. Similar to StyleGANv1v2~\cite{karras2019style,karras2020analyzing} and CIPS~\cite{anokhin2021image}, our model also has the capability of style mixing controlled by stage-wise latent codes as illustrated in Fig.~\ref{fig:style_mixing}.

\clearpage

%
%
\bibliographystyle{splncs04}
\bibliography{egbib}

\end{document}